\documentclass[10pt,twocolumn,letterpaper]{article}

\usepackage{iccv}

\makeatletter
\@namedef{ver@everyshi.sty}{}
\makeatother
\usepackage{tikz}

\usepackage{times}
\usepackage{epsfig}
\usepackage{graphicx}
\usepackage{amsmath}
\usepackage{amssymb}

\usepackage{bm}
\usepackage{rotating}
\usepackage{multirow}
\usepackage{array}
\newcolumntype{?}{!{\vrule width 2pt}}
\usepackage{cite}
\usepackage{tikz}
\usetikzlibrary{shapes.geometric, shapes.misc, positioning, arrows, arrows.meta, decorations.markings}  \usepackage{scalefnt}
\tikzset{every picture/.style={/utils/exec={\sffamily}}}

\usepackage[pagebackref=true,breaklinks=true,letterpaper=true,colorlinks,bookmarks=false]{hyperref}

\usepackage{xr-hyper}
\iccvfinalcopy 

\def\iccvPaperID{4075} 

\newcommand{\PAR}[1]{\vskip4pt \noindent{\bf #1~}}

\definecolor{road}{RGB}{128, 64,128}
\definecolor{sky}{RGB}{70,130,180}
\definecolor{terrain}{RGB}{152,251,152}
\definecolor{bicycle}{RGB}{119, 11, 32}
\definecolor{sign}{RGB}{220,220, 0}
\definecolor{pole}{RGB}{153,153,153}
\definecolor{wall}{RGB}{102,102,156}
\definecolor{cnncolor}{RGB}{200,200,200}
\definecolor{pt1}{RGB}{134,153,67}
\definecolor{pt2}{RGB}{65,127,166}
\definecolor{pt3}{RGB}{168,120,64}
\definecolor{pt4}{RGB}{137,66,162}

\ificcvfinal\fi
\begin{document}

\title{Fine-Grained Segmentation Networks: Self-Supervised Segmentation for Improved Long-Term Visual Localization} 

\author{M\r{a}ns Larsson \hspace{3mm} Erik Stenborg \hspace{3mm} Carl Toft \hspace{3mm} Lars Hammarstrand \hspace{3mm} Torsten Sattler \hspace{3mm} Fredrik Kahl \\
Chalmers University of Technology \\
}

\maketitle
\thispagestyle{empty}

\begin{abstract}
Long-term visual localization is the problem of estimating the camera pose of a given query image in a scene whose appearance changes over time. It is an important problem in practice, for example, encountered in autonomous driving.
In order to gain robustness to such changes, long-term localization approaches often use segmantic segmentations as an invariant scene representation, as the semantic meaning of each scene part should not be affected by seasonal and other changes. However, these representations are typically not very discriminative due to the limited number of available classes. In this paper, we propose a new neural network, the Fine-Grained Segmentation Network (FGSN), that can be used to provide image segmentations with a larger number of labels and can be trained in a self-supervised fashion. In addition, we show how FGSNs can be trained to output consistent labels across seasonal changes. We demonstrate through extensive experiments that integrating the fine-grained segmentations produced by our FGSNs into existing localization algorithms leads to substantial improvements in localization performance.
\end{abstract}

\section{Introduction}
Visual localization is the problem of estimating the camera pose of a given image relative to a visual representation of a known scene. It is a classical problem in computer vision and 
solving the visual localization problem is one key to advanced computer vision applications such as self-driving cars and other autonomous robots, as well as Augmented / Mixed / Virtual Reality. 

\begin{figure}[t]
    \centering
    \setlength\tabcolsep{1pt} 
    \begin{tabular}{ccc}
    Image & Sem. Classes & 100 Clusters \\

    \includegraphics[width=0.15\textwidth]{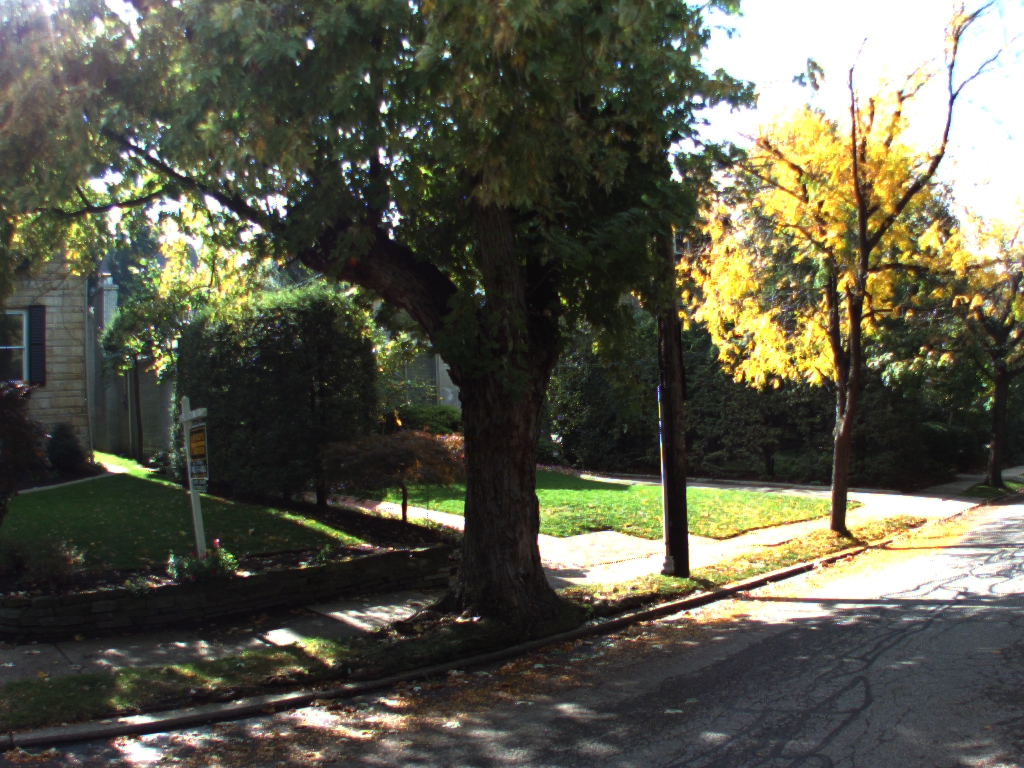} &
    \includegraphics[width=0.15\textwidth]{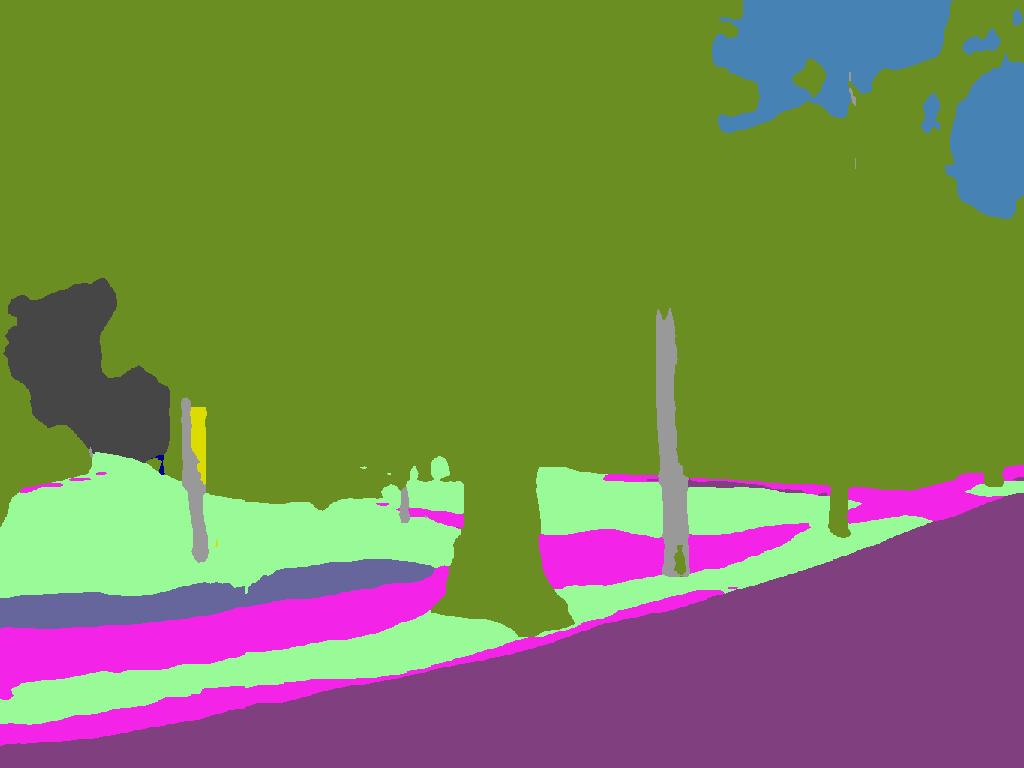} &
    \includegraphics[width=0.15\textwidth]{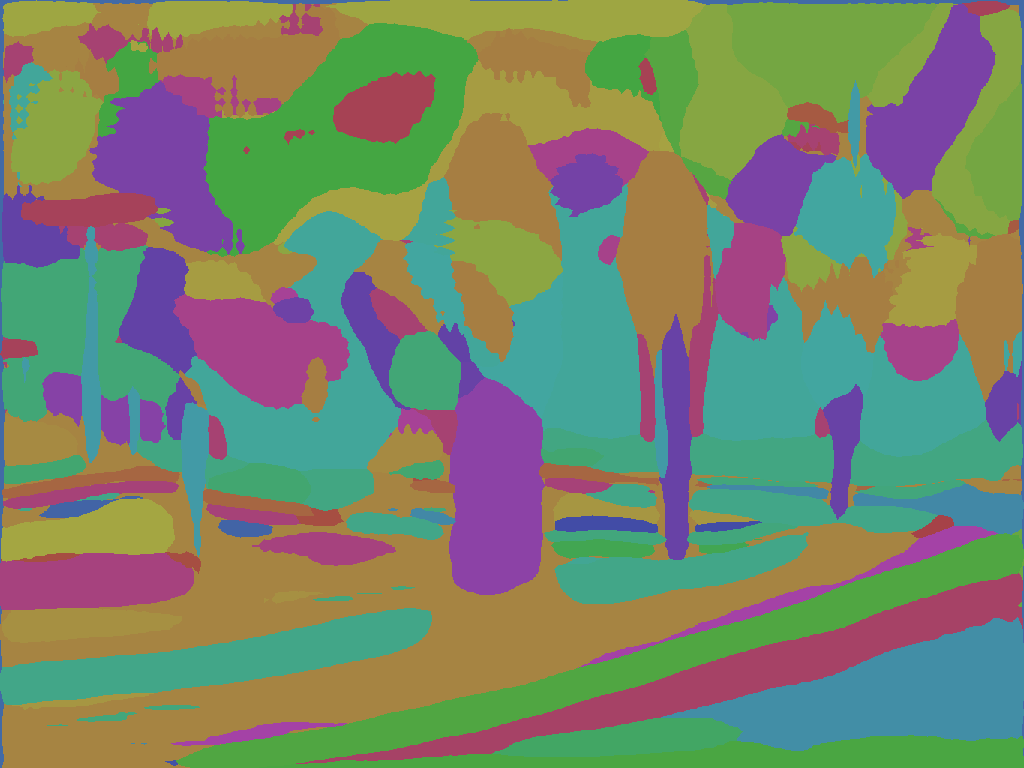} \\
    \end{tabular}
    \setlength\tabcolsep{6pt} 
    \caption{Rather than using a small set of human-defined semantic classes, we train a neural network that automatically discovers a large set of fine-grained clusters. We experimentally show that using a larger number of clusters improves localization performance.  }
    \label{fig:intr}
\end{figure}

The scene representation used by localization algorithms is typically recovered from images depicting a given scene. 
The type of representation can vary from a set of images with associated camera poses~\cite{Sattler2017CVPR,Balntas2018ECCV,Zhang20063DPVT}, over 3D models constructed from Structure-from-Motion~\cite{Schoenberger2016CVPR,Snavely2006SIGGRAPH}, to weights encoded in convolutional neural networks (CNNs)~\cite{Brachmann2017CVPR,Brachmann2018CVPR,Massiceti2017CVPR,Kendall2015ICCV,Kendall2017CVPR,Balntas2018ECCV,Brahmbhatt2018CVPR} or random forests~\cite{Shotton2013CVPR,Cavallari2017CVPR,Brachmann2016CVPR}. 
In practice, capturing a scene from all possible viewpoints and under all potential conditions, \eg, different illumination conditions, is prohibitively expensive~\cite{Sattler2018CVPR}. 
Localization algorithms thus need to be robust to such changes. 

In the context of long-term operation, \eg, under seasonal changes, the scene appearance can vary drastically over time. 
However, the semantic meaning of scene parts remains the same, \eg, a tree is a tree whether it carries leaves or not. 
Based on this insight, approaches for semantic long-term visual localization use semantic segmentations of images or object detections to obtain an invariant scene representation~\cite{Toft2017ICCVW,Toft2018ECCV,Schoenberger2018CVPR,Yu2018IROS,Stenborg2018LongTermVL,Arandjelovic2014ACCV_A,Radwan2018RAL,Cohen2016ECCV,Yu2015CVPR,Singh2016LSVGL,Yu2018IROS}. 
However, this invariance comes at the price of a lower discriminative power as often only few classes are available. 
For example, the Cityscapes dataset~\cite{Cordts2016Cityscapes} uses 19 classes for evaluation, 8 of which cover dynamic objects such as cars or pedestrians that are not useful for localization. 
The Mapillary Vistas dataset~\cite{neuhold2017mapillary} contains 66 classes, with 15 classes for dynamic objects. 
At the same time, annotating more classes comes at significant human labor cost and annotation time. 

In this paper, we show that using significantly more class labels leads to better performance of semantic visual localization algorithms. 
In order to avoid heavy human annotation time, we use the following central insight: 
the image segmentations used by such methods need to be stable under viewpoint, illumination, seasonal, \etc changes. 
However, the classes of the segmentations do not need to map to human-understandable concepts to be useful, \ie., they might not necessarily need to be semantic. 
Inspired by recent work on using $k$-means clustering to pretrain CNNs from unlabelled data~\cite{caron2018deep}, we thus propose a self-supervised, data-driven approach to define fine-grained classes for image segmentation. 
More precisely, we use $k$-means clustering on pixel-level CNN features to define $k$ classes. 
As shown in Fig.~\ref{fig:intr}, this allows our approach, termed Fine-Grained Segmentation Networks (FGSNs), to create more fine-grained segmentations.

In detail, this paper makes the following contributions:
\textbf{1)} We present a novel type of segmentation network, the Fine-Grained Segmentation Network (FGSN), that outputs dense segmentation maps based on cluster indices. This removes the need for human-defined classes and allows us to define classes in a data-driven way through self-supervised learning. Using a 2D-2D correspondence dataset~\cite{larsson2019corr} for training, we ensure that our classes are stable under seasonal and viewpoint changes. 
The source code of our approach is publicly available {\footnote{\url{https://github.com/maunzzz/fine-grained-segmentation-networks}}.
\textbf{2)} FGSNs allow us to create finer segmentations with more classes. We show that this has a positive impact on semantic visual localization algorithms and can lead to substantial improvements when used by existing localization approaches. 
\textbf{3)} We perform detailed experiments to investigate the impact the number of clusters has on multiple visual localization algorithms. In addition, we compare two types of weight initializations, using networks pre-trained for semantic segmentation and image classification, respectively.


\section{Related Work}
The following reviews work related to our approach, most notably semantic segmentation and visual localization.

\PAR{Semantic Segmentation.}
Semantic segmentation is the task of assigning a class label to each pixel in an input image. 
Modern approaches use fully convolutional networks~\cite{long2015fully}, potentially pre-trained for classification~\cite{long2015fully}, while 
incorporating higher level context~\cite{zhao2017pyramid}, enlarging the receptive field~\cite{chen2018deeplab, yu2015multi, Chen_2018_ECCV}, or fusing multi-scale features ~\cite{chen2016attention, ronneberger2015u}. Another line of work combines FCNs with probabilistic graphical models, \eg, in the form of a post-processing step~\cite{chen2018deeplab} or as a differentiable component in an end-to-end trainable network~\cite{zheng2015conditional, liu2015semantic, larsson2018revisiting}.

CNNs for semantic segmentation are usually trained in a fully supervised fashion.  
However, obtaining a large amount of densely labeled images is very time-consuming and expensive~\cite{Cordts2016Cityscapes, neuhold2017mapillary}. 
As a result, approaches based on weaker forms of annotations have been developed. Some examples of weak labels used to train FCNs are bounding boxes~\cite{dai2015boxsup, khoreva2017simple, papandreou2015weakly}, image level tags~\cite{papandreou2015weakly, pathak2015constrained, Pinheiro_2015_CVPR, souly2017semi}, points~\cite{bearman2016s}, or 2D-2D point matches~\cite{larsson2019corr}. 
In this paper, we show that the classes used for ``semantic" visual localization do not need to carry semantic meaning. 
This allows us to directly learn a large set of classes for image segmentation from data in a self-supervised fashion. 
During training, we use 2D-2D point matches~\cite{larsson2019corr} to encourage consistency of the segmentations across seasonal changes and across different weather conditions.

\PAR{(Semantic) Visual Localization.}
Traditionally, approaches for visual localization use a 3D scene model constructed from a set of database images via Structure-from-Motion~\cite{Li2012ECCV,Sattler2017PAMI,Cao14CVPR,Choudhary12ECCV,Lim2015IJRR,larsson-etal-bmvc-2016,Svarm2017PAMI,Zeisl2015ICCV}. 
Associating each 3D model point with local image features such as SIFT~\cite{Lowe04IJCV}, these approaches establish a set of 2D-3D correspondences between a query image and the model via descriptor matching. 
The resulting matches are then used for RANSAC-based camera pose estimation~\cite{Fischler81CACM}. 
Machine learning-based approaches either replace the 2D-3D matching stage through scene coordinate regression~\cite{Brachmann2017CVPR,Brachmann2018CVPR,Cavallari2017CVPR,Massiceti2017CVPR,Shotton2013CVPR,Meng2017IROS,Meng2018IROS}, \ie, they regress the 3D point coordinate in each 2D-3D match, or directly regress the camera pose from an image~\cite{Kendall2015ICCV,Kendall2017CVPR,Brahmbhatt2018CVPR,Walch2017ICCV,Balntas2018ECCV}. 
The former type of methods achieves state-of-the-art localization accuracy in small-scale scenes~\cite{Brachmann2018CVPR,Cavallari2017CVPR,Meng2017IROS}, but do not seem to easily scale to larger scenes~\cite{Brachmann2018CVPR}. 
The latter type of methods have recently been shown to not perform consistently better than image retrieval methods~\cite{Sattler2019CVPR}, \ie, approaches that approximate the pose of the query image by the pose of the most similar database image~\cite{Torii15CVPR,Arandjelovic16CVPR,Kim2017CVPR}. 
As such, state-of-the-art methods for long-term visual localization at scale either rely on local features for matching~\cite{Sarlin2019CVPR,Schoenberger2018CVPR,Toft2018ECCV,Toft2017ICCVW,Stenborg2018LongTermVL,Garg2018RSS} or use image retrieval techniques~\cite{Torii15CVPR,Arandjelovic16CVPR,Anoosheh2019ICRA,Porav2018ICRA,Arandjelovic2014ACCV_A,Singh2016LSVGL,Yu2018IROS}. 

One class of semantic visual localization approaches uses object detections as features~\cite{Atanasov2016IJRR,Ardeshir2014ECCV,Salas-Moreno2013CVPR}. 
In this paper, we focus on a second class of approaches based on semantic segmentations~\cite{Schoenberger2018CVPR,Toft2018ECCV,Toft2017ICCVW,Stenborg2018LongTermVL,Garg2018RSS,Cohen2016ECCV,Arandjelovic2014ACCV_A,Singh2016LSVGL,Yu2018IROS}. 
These methods use semantic image segmentations to obtain a scene representation that is invariant to appearance and (moderate) geometry changes. 
Due to the small number of classes typically available, the resulting representation is not very discriminative. 
Thus, semantic localization approaches use semantics as a second sensing modality next to 3D information~\cite{Schoenberger2018CVPR,Cohen2016ECCV,Toft2018ECCV,Toft2017ICCVW,Stenborg2018LongTermVL}. 
In this paper, we show that the image segmentations used by such methods do not necessarily need to be semantic. 
Rather, we show that these approaches benefit from the more fine-grained segmentations with more classes produced by our FGSNs.

\PAR{Domain Adaption.} 
Semantic localization algorithms implicitly assume that  semantic segmentations are robust to illumination, viewpoint, seasonal, and other changes. 
In practice, CNNs for semantic segmentation typically only perform well under varying conditions if these conditions are reflected in the training set. 
Yet, creating pixel-level annotations for large image sets is a time consuming task. 
Domain adaptation approaches~\cite{kulis2011you, saenko20adapting, ganin2016domain, long2015learning, long2016unsupervised, tzeng2017adversarial, zhu2017unpaired} thus consider the problem of applying algorithms trained on one domain to new domains, where little to no labeled data is available. 
This makes training on synthetic datasets~\cite{ros2016synthia,richter2016playing} to improve the performance on real images~\cite{hoffman2016fcns, zou2018unsupervised, sankaranarayanan2018learning} possible. In addition, the performance of the network on images taken during different weather and lighting conditions can be improved~\cite{wulfmeier2017addressing, wulfmeier2018incremental}. 
In the context of (semantic) image segmentation, these approaches improve the robustness of the segmentations. 
However, they do not increase the number of available classes and are thus complimentary to our approach. 
We use a recently proposed correspondence dataset~\cite{larsson2019corr} for the same purpose, to ensure that our segmentations are robust to illumination and seasonal changes. 

\PAR{Self-Supervised Learning.} Self-supervised learning approaches are a variant of unsupervised learning methods, where a model learns to predict a set of  labels that can be automatically created from the input data. Several approaches train a CNN to perform a domain specific auxiliary task~\cite{doersch2015unsupervised, paulin2015local, zhang2016colorful, noroozi2017representation}. Some examples of tasks include predicting missing image parts~\cite{Pathak_2016_CVPR}, ego motion~\cite{Agrawal_2015_ICCV}, and the rotation of an image~\cite{gidaris2018unsupervised}. To solve these auxiliary tasks, the CNNs need to learn meaningful visual features that can then also be used for the actual task at hand. In \cite{caron2018deep}, Caron \etal train a CNN for the task of image-level classification using labels acquired by $k$-means clustering of image features. We extend this approach to training an image segmentation network. We also use the actual clusters, or labels, explicitly for visual localization. This in contrast to ~\cite{caron2018deep}, where the clusters are just a means for learning features for tasks such as classification.

\tikzstyle{cnn} = [trapezium, rotate=-90, trapezium left angle=70, trapezium right angle=70, rounded corners, minimum width=1, minimum height=3,text centered, draw=black, fill=cnncolor, ultra thick]
\tikzstyle{pre}=[<-,shorten <=10pt, shorten >=10pt, ultra thick]
\tikzstyle{post}=[->,shorten <=10pt ,shorten >=10pt, ultra thick]
\tikzset{cross/.style={cross out, draw, 
         minimum size=2*(#1-\pgflinewidth), 
         inner sep=0pt, outer sep=0pt, ultra thick}}
         
\begin{figure*}
\centering
    
\resizebox{\textwidth}{!}{
\begin{tikzpicture}

\pgfmathsetmacro{\xmidlabel}{-7}
\pgfmathsetmacro{\xmidtrain}{13}

\draw[black, ultra thick, rounded corners] (\xmidlabel-7.7, -4.5) rectangle (\xmidlabel+10.1, 4.5);
\node at (\xmidlabel-5, 5.1) {\huge Label creation};

\draw[black, ultra thick, rounded corners] (\xmidtrain-6.7, -4.5) rectangle (\xmidtrain+11.2, 4.5);
\node at (\xmidtrain-5, 5.1) {\huge Training};

\draw [->, ultra thick] (\xmidlabel+10.4, .5) to [bend left=70](\xmidtrain-7.0, .5);
\draw [<-, ultra thick] (\xmidlabel+10.4, -.5) to [bend right=70](\xmidtrain-7.0, -.5);

\begin{scope}[shift={(\xmidlabel,0)}]
\node at (-5.2, 3.95) {\large \begin{tabular}{c} Reference \\ Images \end{tabular}};
\node (im1) at (-5.2, 2) {\includegraphics[width=3.5cm]{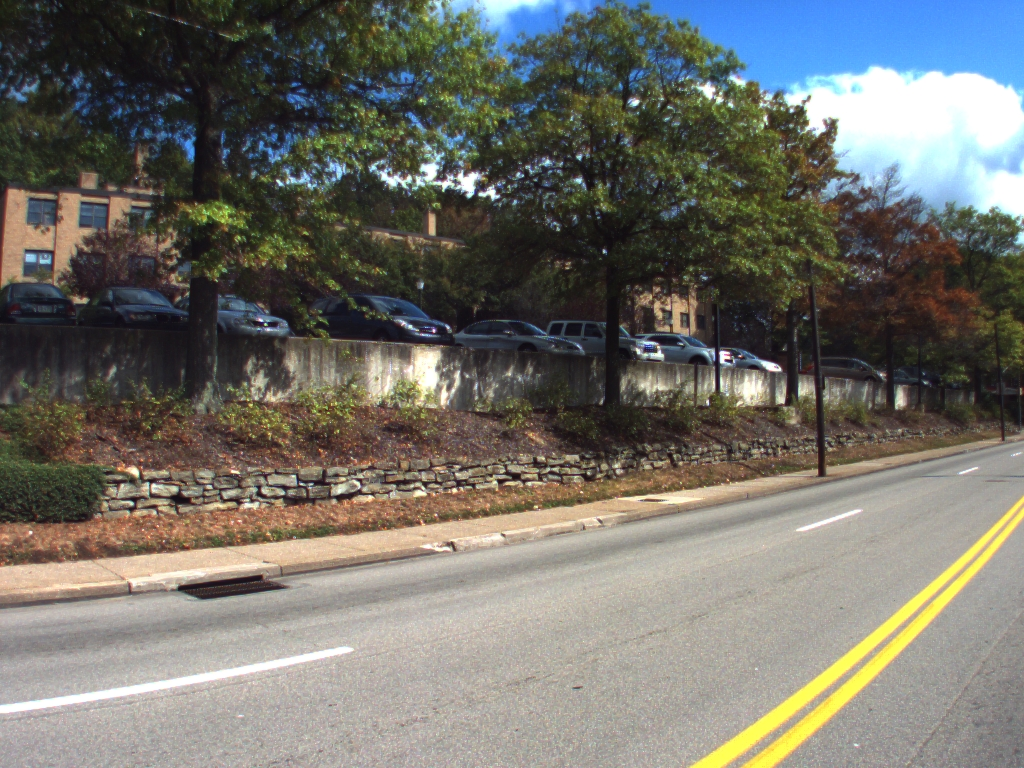}};
\node (im2) at (-5.2, -2)
{\includegraphics[width=3.5cm]{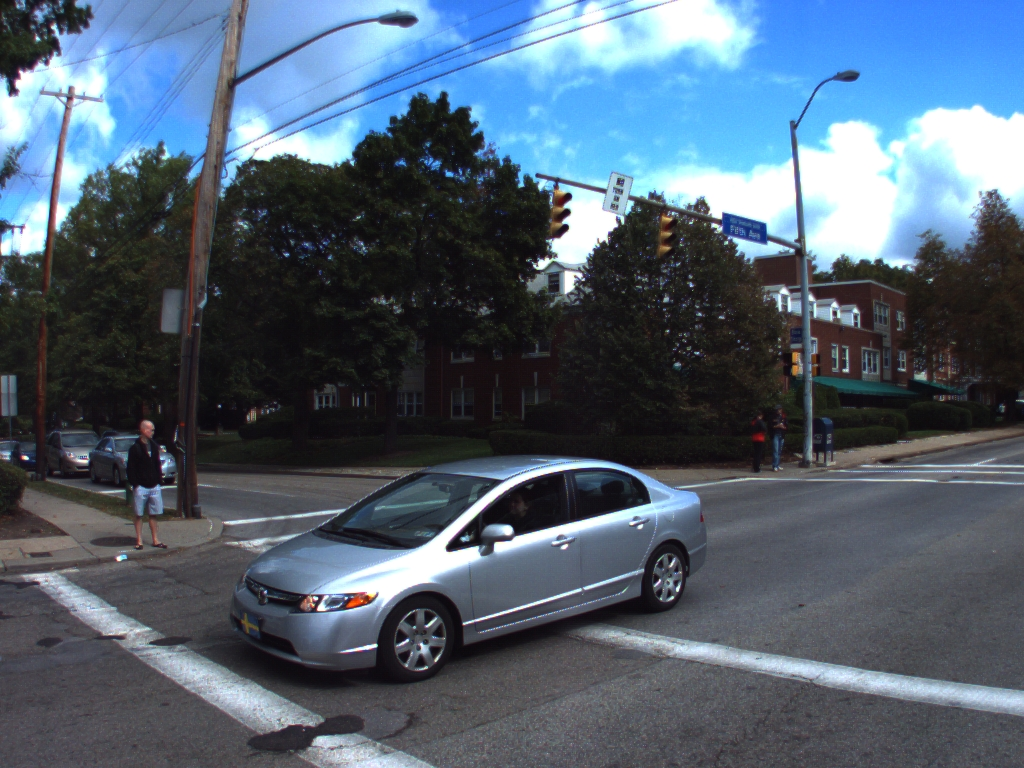}};


\node[cnn] (cnn1) at (-1.9, 2) {\rotatebox{90}{\LARGE FGSN}};
\node[cnn] (cnn2) at (-1.9, -2) {\rotatebox{90}{\LARGE FGSN}};

\node at (-5.2, -.25)[circle,fill,inner sep=1.5pt]{};
\node at (-5.2, 0)[circle,fill,inner sep=1.5pt]{};
\node at (-5.2, .25)[circle,fill,inner sep=1.5pt]{};
\node at (-1.9, -.25)[circle,fill,inner sep=1.5pt]{};
\node at (-1.9, 0)[circle,fill,inner sep=1.5pt]{};
\node at (-1.9, .25)[circle,fill,inner sep=1.5pt]{};
\node at (.5, -.25)[circle,fill,inner sep=1.5pt]{};
\node at (.5, 0)[circle,fill,inner sep=1.5pt]{};
\node at (.5, .25)[circle,fill,inner sep=1.5pt]{};
\node at (7.6, -.25)[circle,fill,inner sep=1.5pt]{};
\node at (7.6, 0)[circle,fill,inner sep=1.5pt]{};
\node at (7.6, .25)[circle,fill,inner sep=1.5pt]{};

\pgfmathsetmacro{\cubex}{.7}
\pgfmathsetmacro{\cubey}{1.4}
\pgfmathsetmacro{\cubez}{1.8}
\pgfmathsetmacro{\xposa}{.5}
\pgfmathsetmacro{\yposa}{2.5}
\pgfmathsetmacro{\xposb}{.5}
\pgfmathsetmacro{\yposb}{-1.5}
\pgfmathsetmacro{\op}{.6}
\pgfmathsetmacro{\nsq}{5}


\node at (0.5*\xposa+0.5*\xposb + 3, 2.5) {\large \begin{tabular}{c} k-means \\ clustering \end{tabular}};

\node (schem) at (0.5*\xposa+0.5*\xposb + 3, 0) {\includegraphics[width=3.5cm]{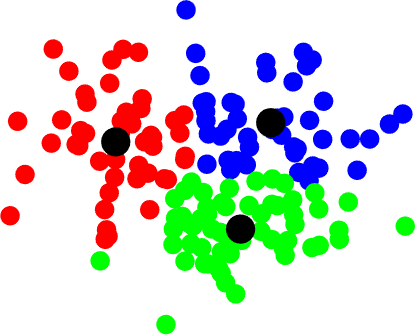}};

\pgfmathsetmacro{\alen}{.75}
\draw [->, ultra thick] (0.5*\xposa+0.5*\xposb+1, 2.2) -- ++(\alen,-\alen);
\draw [->, ultra thick] (0.5*\xposa+0.5*\xposb+1, -1.7) -- ++(\alen,\alen);

\draw [<-, ultra thick] (7.6-3.5/2-.5, 2.2) -- ++(-\alen,-\alen);
\draw [<-, ultra thick] (7.6-3.5/2-.5, -1.7) -- ++(-\alen,\alen);

\foreach \i in {-4,...,0}
{
    \foreach \j in {-4,...,0}
    {
        \draw[black, fill=sky] (\xposa,\yposa + \i * \cubey / \nsq,\j * \cubez / \nsq) -- ++(-\cubex ,0,0) -- ++(0,-\cubey / \nsq,0) -- ++(\cubex,0,0) -- cycle;
        \draw[black, fill=sky] (\xposa,\yposa + \i * \cubey / \nsq, \j * \cubez / \nsq) -- ++(0,0,-\cubez / \nsq) -- ++(0,-\cubey / \nsq,0) -- ++(0,0,\cubez / \nsq) -- cycle;
        \draw[black, fill=sky] (\xposa,\yposa + \i * \cubey / \nsq, \j * \cubez / \nsq) -- ++(-\cubex,0,0) -- ++(0,0,-\cubez / \nsq) -- ++(\cubex,0,0) -- cycle;
    }        
}
\draw[fill=sky, opacity=\op] (\xposa,\yposa,0) -- ++(-\cubex,0,0) -- ++(0,-\cubey,0) -- ++(\cubex,0,0) -- cycle;
\draw[fill=sky, opacity=\op] (\xposa,\yposa,0) -- ++(0,0,-\cubez) -- ++(0,-\cubey,0) -- ++(0,0,\cubez) -- cycle;
\draw[fill=sky, opacity=\op] (\xposa,\yposa,0) -- ++(-\cubex,0,0) -- ++(0,0,-\cubez) -- ++(\cubex,0,0) -- cycle;

\foreach \i in {-4,...,0}
{
    \foreach \j in {-4,...,0}
    {
        \draw[black, fill=terrain] (\xposb,\yposb + \i * \cubey / \nsq,\j * \cubez / \nsq) -- ++(-\cubex ,0,0) -- ++(0,-\cubey / \nsq,0) -- ++(\cubex,0,0) -- cycle;
        \draw[black, fill=terrain] (\xposb,\yposb + \i * \cubey / \nsq, \j * \cubez / \nsq) -- ++(0,0,-\cubez / \nsq) -- ++(0,-\cubey / \nsq,0) -- ++(0,0,\cubez / \nsq) -- cycle;
        \draw[black, fill=terrain] (\xposb,\yposb + \i * \cubey / \nsq, \j * \cubez / \nsq) -- ++(-\cubex,0,0) -- ++(0,0,-\cubez / \nsq) -- ++(\cubex,0,0) -- cycle;
    }        
}

\draw[black,fill=terrain, opacity=\op] (\xposb,\yposb,0) -- ++(-\cubex,0,0) -- ++(0,-\cubey,0) -- ++(\cubex,0,0) -- cycle;
\draw[black,fill=terrain, opacity=\op] (\xposb,\yposb,0) -- ++(0,0,-\cubez) -- ++(0,-\cubey,0) -- ++(0,0,\cubez) -- cycle;
\draw[black,fill=terrain, opacity=\op] (\xposb,\yposb,0) -- ++(-\cubex,0,0) -- ++(0,0,-\cubez) -- ++(\cubex,0,0) -- cycle;

\node at (\xposa, 3.9) {\large Features};

\node at (7.6, 3.9) {\large Labels};
\node (cl1) at (7.6, 2) {\includegraphics[width=3.5cm]{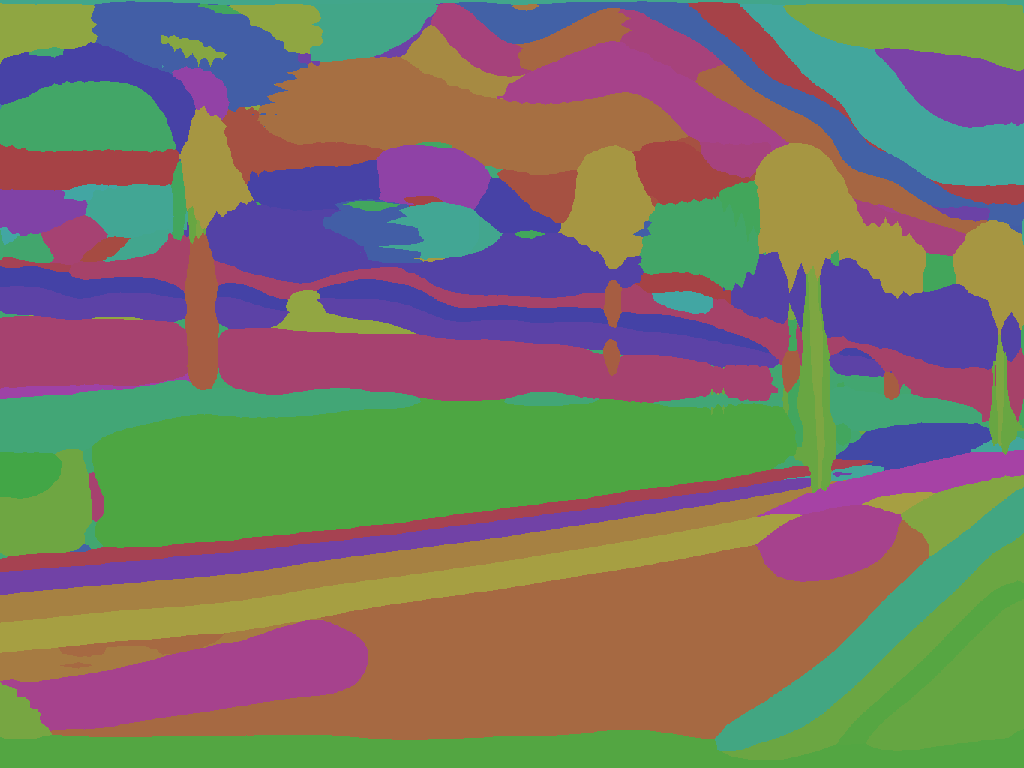}};
\node (cl2) at (7.6, -2) {\includegraphics[width=3.5cm]{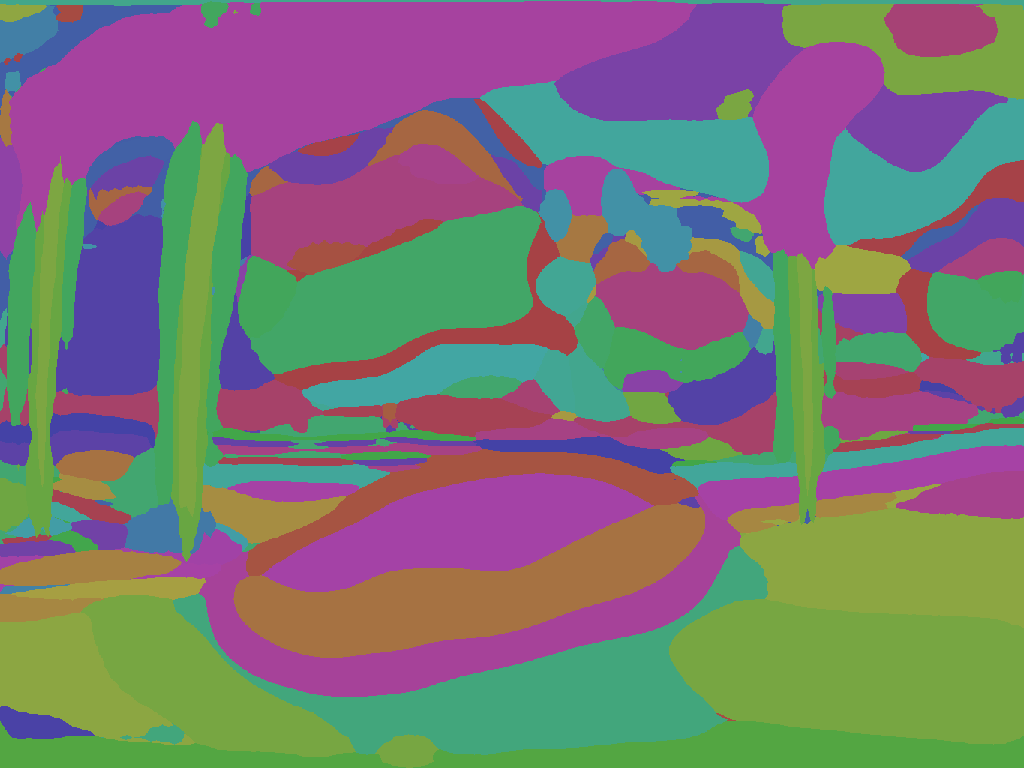}};
\end{scope}

\begin{scope}[shift={(\xmidtrain,0)}]

\node at (-4.3, 3.93) {\large Reference};
\node at (-4.3, -.3) {\large Target};
\node (im1) at (-4.3, 2) {\includegraphics[width=3.5cm]{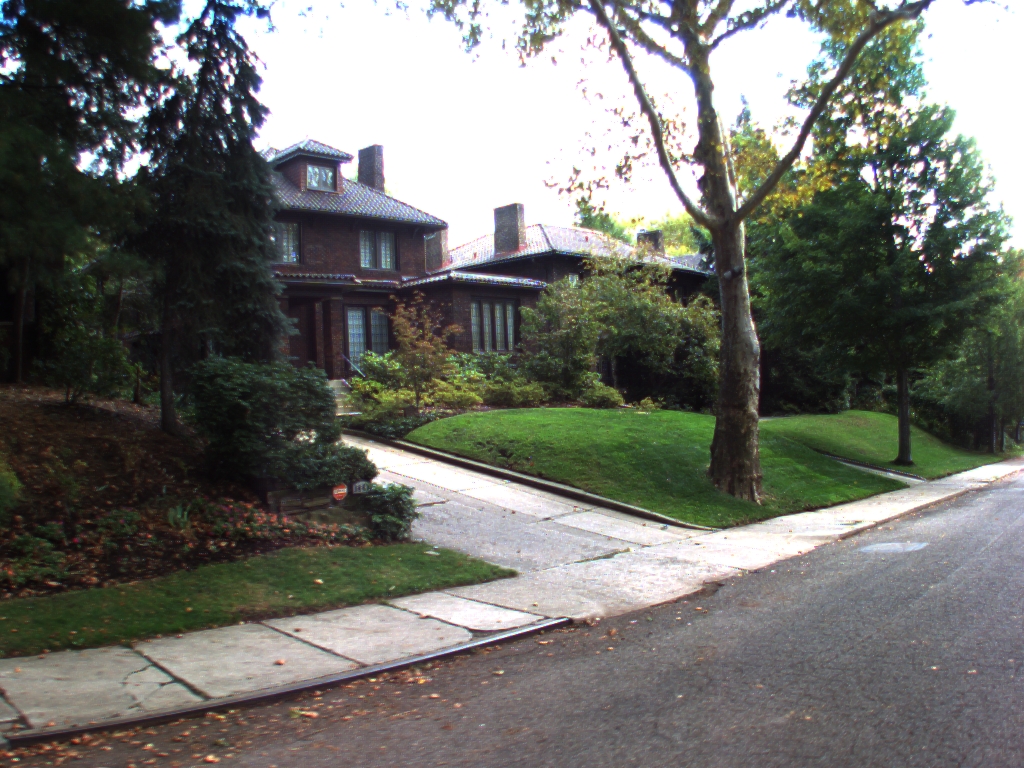}};
\node (im2) at (-4.3, -2) {\includegraphics[width=3.5cm]{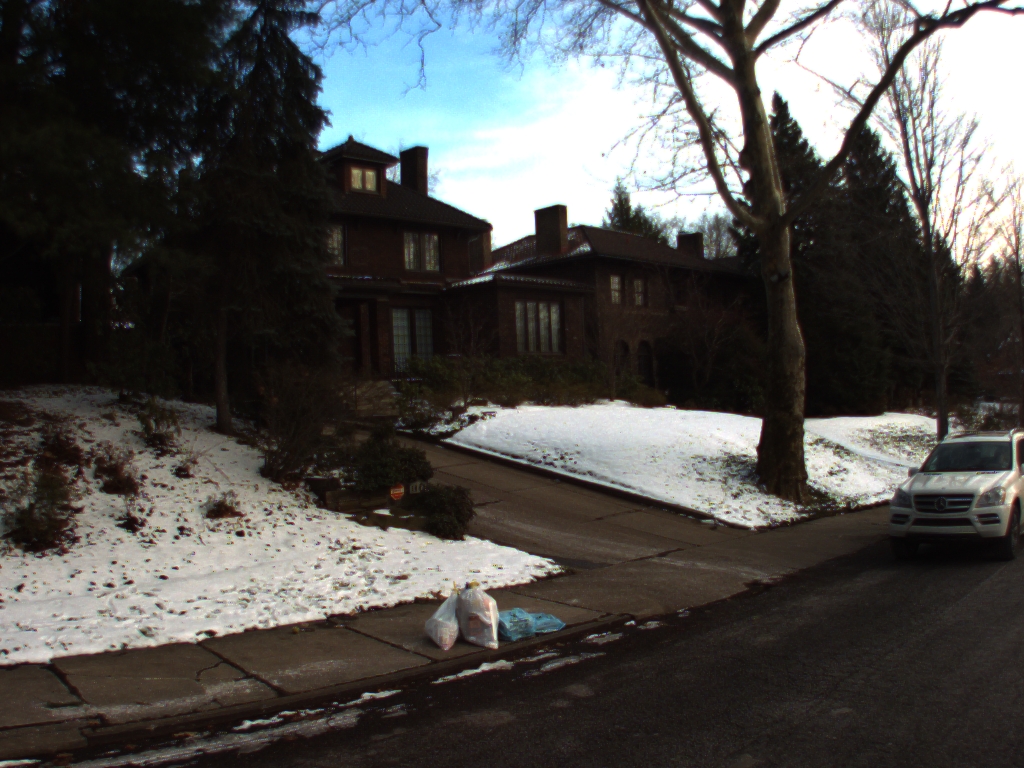}};

\node[cnn] (cnn1) at (-1, 2) {\rotatebox{90}{\LARGE FGSN}};
\node[cnn] (cnn2) at (-1, -2) {\rotatebox{90}{\LARGE FGSN}}
    edge [pre ,bend left=45]  (cnn1)
    edge [post ,bend right=45] (cnn1);
\node (ws) at (-1, 0) {\parbox{2cm}{\centering \large Weight Sharing} };    

\draw[->,line width=1mm,>=latex, text opacity=1] (.4, 2 + .2) -- node [above] {\normalsize Forward} ++(1.7 , 0);
\draw[->,line width=1mm,>=latex, text opacity=1] (.4, -2 + .2) -- node [above] {\normalsize Forward} ++(1.7 , 0);

\draw[<-,line width=1mm, dashed,>=latex, text opacity=1] (.4, 2 - .2) -- node [below] {\normalsize Backward} ++(1.7 , 0);
\draw[<-,line width=1mm, dashed,>=latex, text opacity=1] (.4, -2 - .2) -- node [below] {\normalsize Backward} ++(1.7 , 0);

\node (im1) at (4.2, 2) {\includegraphics[width=3.5cm]{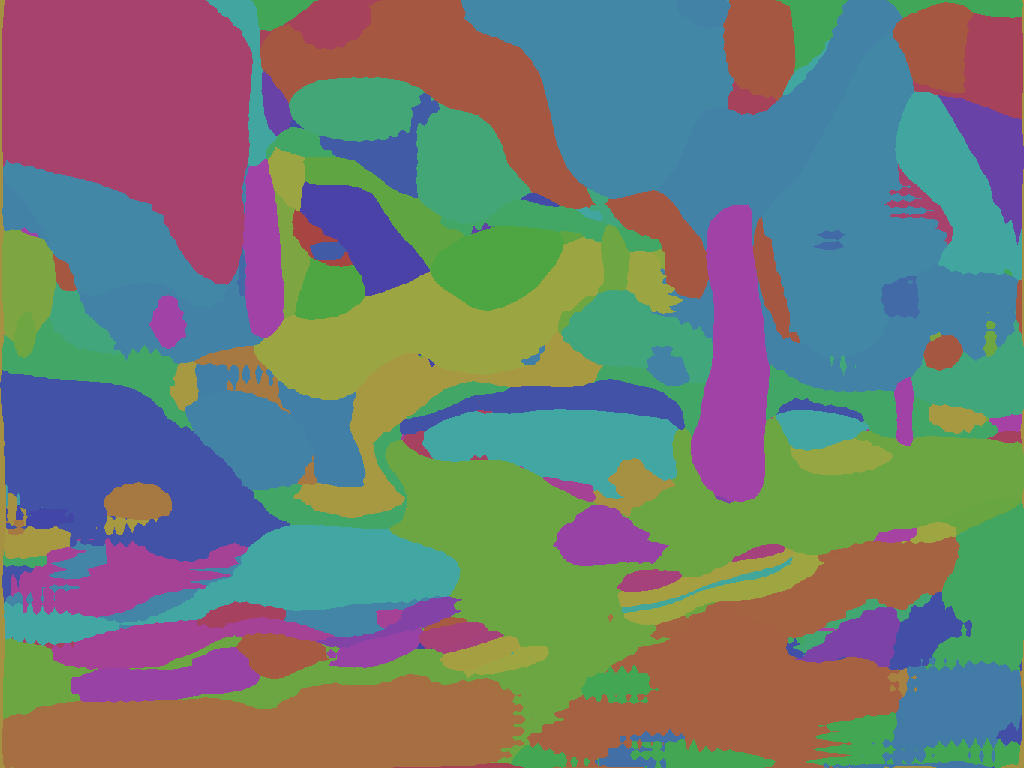}};
\node (im2) at (4.2, -2) {\includegraphics[width=3.5cm]{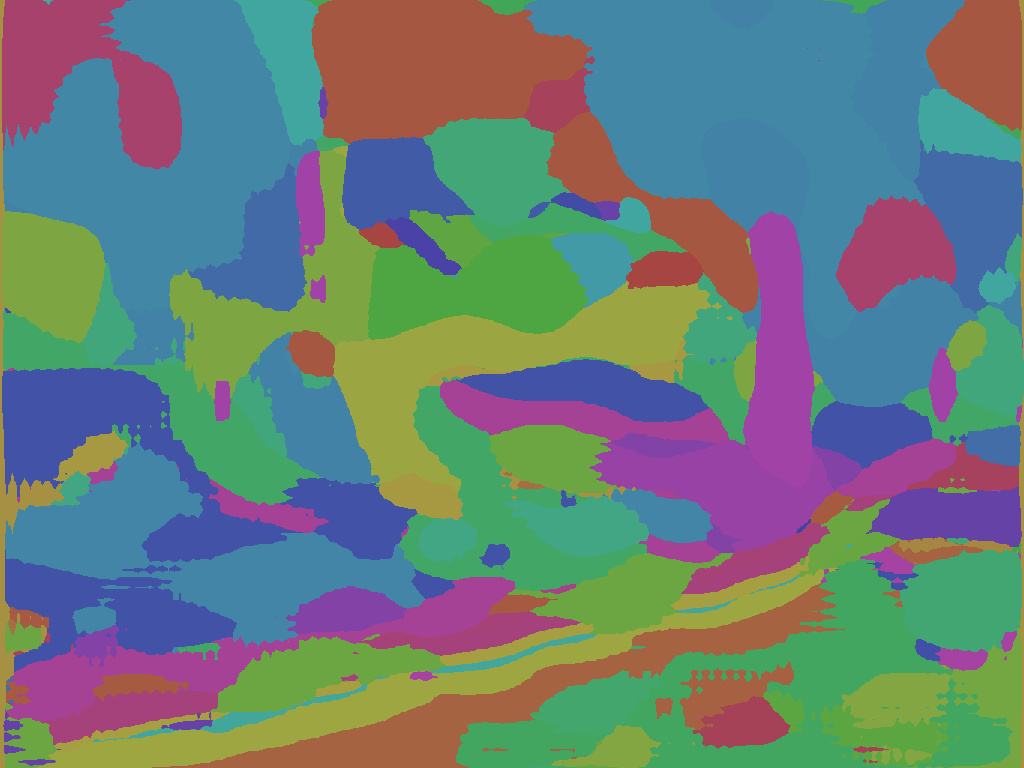}};

\node (im1) at (8.3, 2) {\includegraphics[width=3.5cm]{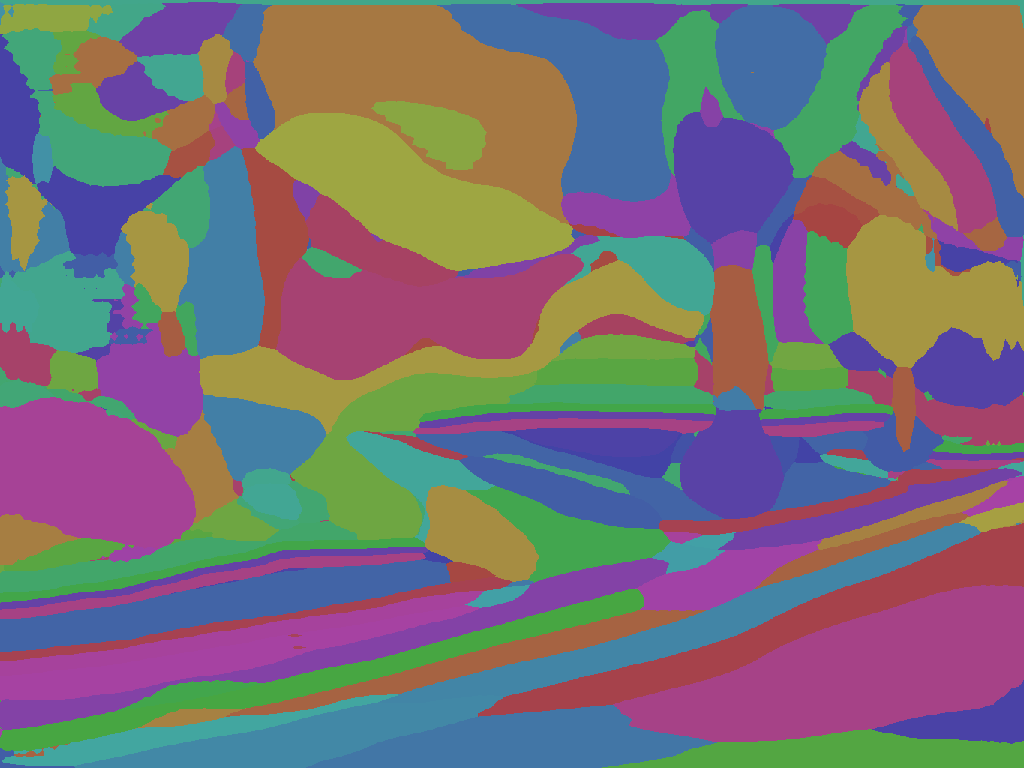}};
\draw[black, ultra thick] (8.3-3.5/2, -3.3) rectangle (8.3+3.5/2, -.7);

\foreach \x/\yshift in {6.7cm/0cm,7.4cm/-.1cm,8.4cm/1cm,9cm/.5cm}
    {\draw[ultra thick] (\x , 2cm+\yshift) to [bend left](\x , -3cm+\yshift);
    \draw (\x, 2cm+\yshift) node[cross=7pt,sign]{};
    \draw (\x, -3cm+\yshift) node[]{};}

\node at (6.7cm, -3cm)[circle,fill=pt1,inner sep=2.5pt]{};
\node at (7.4cm, -3.1cm)[circle,fill=pt2,inner sep=2.5pt]{};
\node at (8.4, -2cm)[circle,fill=pt3,inner sep=2.5pt]{};
\node at (9, -2.5cm)[circle,fill=pt4,inner sep=2.5pt]{};
    

\draw[black, dashed, ultra thick, rounded corners] (4.2-3.9/2, -3.5) rectangle (8.3+3.9/2, -.5);
\draw[black, dashed, ultra thick, rounded corners] (4.2-3.9/2, .5) rectangle (8.3+3.9/2, 3.5);    

\node at (10.7, 2) {\rotatebox{270}{\LARGE $\mathcal{L}_{class}$}};
\node at (4.2, 3.93) {\large Output};
\node at (10.7, -2) {\rotatebox{270}{\LARGE $\mathcal{L}_{corr}$}};
\node at (8.3, 3.98) {\large \begin{tabular}{c} Labels and \\ Correspondences \end{tabular}};

\end{scope}
\end{tikzpicture}
}
\caption{Illustration of the training procedure of an FGSN. To create the training data, features are extracted from all reference images from the correspondence dataset. The features are then clustered using $k$-means clustering and the assignments are used as labels for the images. In addition to having dense labels for the reference images, we also use the 2D-2D correspondences during training to encourage consistency across weather conditions and seasons as well as varying viewpoints.}
\label{fig:cluster_network_training}
\end{figure*}
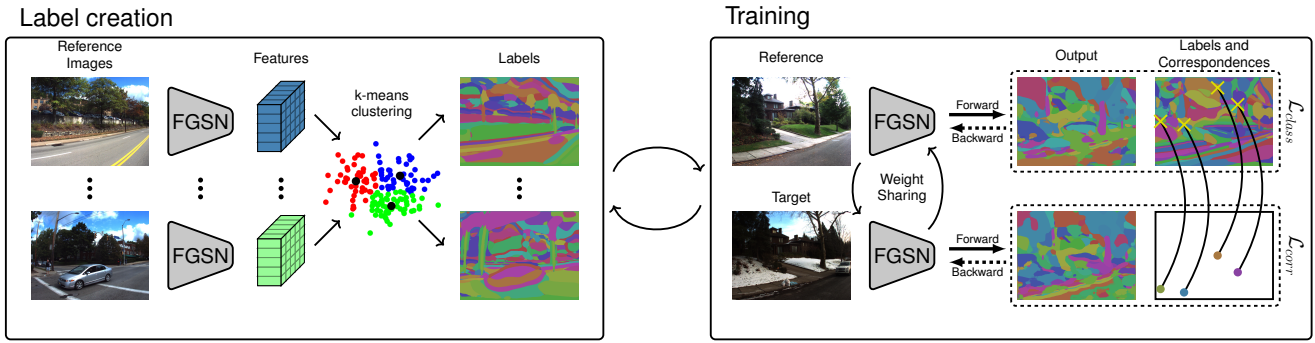

\section{Fine-Grained Segmentation Networks}
The Fine-Grained Segmentation Network (FGSN) has the same structure as a standard CNN used for semantic segmentation. Given an input image, it produces a dense segmentation map. However, instead of being trained on a set of manually created annotations, labels are created in a self-supervised manner. During training, at certain intervals, features are extracted from the images in the training set and clustered using $k$-means clustering. The cluster assignments, one at each pixel, are then used as supervision during training, \ie as labels. In this way, we can change the number of classes that the FGSN outputs without having to create annotations with the new set of classes. The FGSN is trained to output the correct label for each pixel. 

We also use a set of 2D-2D point correspondences~\cite{larsson2019corr} during training to ensure that the predictions are stable under seasonal changes and viewpoint variations. Each sample of the correspondence dataset contains two images of the same scene taken from different traversals and thus in different seasonal or weather conditions. One of the images in each pair is always from a  the reference traversal, captured during favourable weather conditions. A set of 2D-2D point correspondences between points in the images depicting the same 3D point is also available for each image pair. The network is encouraged to predict the same class for the two points in each correspondence to make the output robust to seasonal changes. Fig.~\ref{fig:cluster_network_training} illustrates the training process. Note that creating the correspondence dataset is a significantly less laborious process than hand-labeling the same images with semantic labels, see details in~\cite{larsson2019corr}.


\PAR{Label creation}
For the creation of the labels we use the method developed by Caron \etal~\cite{caron2018deep} based on $k$-means clustering. We, however, need to do some modifications to make it work well for dense output and training with 2D-2D correspondences. The main idea is to do $k$-means clustering on the output features of the CNN, then add a layer to the network and train using the cluster assignments as labels. After a fixed number of training iterations, the clustering is repeated and the final layer re-initialized. 

For clustering we extract features from all images in the reference traversal of the correspondence dataset. This traversal contains images captured during favourable weather conditions, hence if we initialize the network with weights trained for semantic segmentation the features extracted will contain meaningful semantic information.
%
For each image we get a dense map of image features, from which we randomly sample a set of features for clustering.
Half of the features are extracted from pixel positions where we have 2D-2D correspondences and half are randomly sampled across the entire image. Given the set of extracted image features, clustering is done by solving
\begin{align} 
  \min_{C \in \mathbb{R}^{d\times m}}
  & \frac{1}{N}
  \sum_{n=1}^N
  \min_{\bm{y}_n \in \{0,1\}^{m}}
  \| \bm{d}_n -  C \bm{y}_n \|_2^2
  \\ 
  &\text{s. t.}
  \quad
  \bm{y}_n^\top \bm{1}_m = 1, \nonumber
\end{align}
where $\bm{d}_n$ are feature vectors of length m sampled from the output feature maps produced by the CNN. Solving this problem provides a centroid matrix $C^*$ and a set of optimal assignments $(\bm{y}_n^*)$. To avoid trivial solutions with empty clusters we do a reassignment of the centroids of empty clusters. For each empty cluster centroid, a centroid of a non-empty cluster is randomly choosen. The centroid of the empty cluster is then set to the same value as this centroid with a small perturbation~\cite{caron2018deep, JDH17}. 

\PAR{Training Loss.}
Our training loss consists of two parts, a correspondence part $\mathcal{L}_{corr}$ and a cluster classification part $\mathcal{L}_{class}$. The latter encourages the model to output the correct label for each pixel in the reference images of dataset. We use a standard cross-entropy loss with the labels as targets. The final $\mathcal{L}_{class}$ loss is an average of over all samples.

For $\mathcal{L}_{corr}$, we use the 2D-2D point correspondences. Denote the content of one sample from the correspondence dataset as $(I^{r}, I^{t}, \bm{x}^r, \bm{x}^t)$. Here $I^{r}$ is an image from the reference traversal, $I^{t}$ is an image from the target traversal\footnote{We refer to the second traversal as the target as we aim to ensure that its labeling is consistent with the reference traversal.}, and $\bm{x}^r$ as well as $\bm{x}^t$ are the pixel positions of the matched points in the reference and target images, respectively. 

The correspondence loss function $\mathcal{L}_{corr}$ is an average over all such samples
\begin{equation}
\mathcal{L}_{corr}
= \frac{1}{M}\sum_{(r,t)} l_{CE}(I^{r}, I^{t}, \bm{x}^r, \bm{x}^t) \enspace,
        \label{eq:lossloss}
\end{equation}
where M is the number of samples and $l_{CE}$ is the cluster correspondence cross-entropy loss. Let $\bm{d}_x \in \mathbb{R}^C$ denote the output feature vector of the network of length $C$, the number of clusters, at pixel position $x$. To calculate $l_{CE}$ we begin by taking the cluster assignments, \ie the labels, for the features in the reference image for all positions $\bm{x}^r$. By describing the  label for a pixel at position $x_i$ using the one-hot encoding vector $\bm{c}_{x_i}$, the loss can be written as
\begin{equation}
    l_{CE} = -\frac{1}{N}\sum_{i=1}^N \bm{c}^T_{x^r_i} \left ( \log  ( \bm{d}_{x^r_i}  ) + \log  ( \bm{d}_{x^t_i}  ) \right ) \enspace,  
    \label{eq:CE}
\end{equation}
where $\log(\cdot)$ is taken element-wise. The loss will encourage the pixels in the target image to have the same labels as the corresponding pixels in the reference image. 

%

During training, we minimize $\mathcal{L} = \mathcal{L}_{class} + \mathcal{L}_{corr}$.
%

\PAR{Implementation Details.} 
During the training of the CNN, we minimize the loss $\mathcal{L}$ using stochastic gradient descent with momentum and weight decay. During all experiments the learning rate was set to $2.5 \cdot 10^{-5}$, while the momentum and weight decay were set to $0.9$ and $ 10^{-4}$, respectively. We used the PSPNet~\cite{zhao2017pyramid} network structure with a Resnet101~\cite{he2016deep} base. Due to GPU memory limitations we train with a batch size of one. The networks are trained for 60000 iterations and use the weights that obtained the lowest correspondence loss $\mathcal{L}_{corr}$ on the validation set. The training and evaluation are implemented in PyTorch~\cite{paszke2017automatic}.

After every 10000 iterations, a new set of image features are extracted from the reference images and a new set of cluster centroids and labels are calculated. The features, of initial dimension $512$, are PCA-reduced to $256$ dimensions, whitened and $l_2$-normalized. The $k$-means clustering was done using the Faiss framework~\cite{JDH17}. After clustering, the final layer of the network is randomly re-initialized using a normal distribution with mean $0$ and standard deviation $0.01$. The bias weights are all set to $0$. 


All evaluation and testing is done in patches of size $713 \times 713$ pixels on the original image scale only. Patches are extracted from the image with a step size of $476$ pixels in both directions. The network output is paired with an interpolation weight map that is $1$ for the $236 \times 236$ center pixels of the patch and drops off linearly to $0$ at the edges. For each pixel the weighted mean, using the interpolation maps as weights, is used to produce the pixel's class scores. The motivation behind the interpolation is that the network generally performs better at the center of the patches, since there is more information about the surroundings available. 

\section{Semantic Visual Localization}
\label{sec:localization_baselines}
This paper was motivated by the hypotheses that being able to obtain more fine-grained image segmentations will have a positive impact on semantic visual localization approaches. 
To test this hypothesis, we integrate the segmentations obtained with our FGSNs into multiple semantic visual localization algorithms. 
In the following, we briefly review these algorithms. 
All of them assume that a 3D point cloud of the scene, where each 3D point is associated with a class or cluster label, is available. 
Since the point clouds are linked to images, the labels are obtained by projecting the segmentations of the images onto the point cloud.

\PAR{Simple Semantic Match Consistency (SSMC)~\cite{Toft2018ECCV}.} 
The first approach is a simple-to-implement match consistency filter used as a baseline method in~\cite{Toft2018ECCV}. 
Given a set of 2D-3D matches between features in a query image and 3D points in a Structure-from-Motion (SfM) point cloud, SSMC uses semantics to filter out inconsistent matches. 
A match between a feature $f$ and a 3D point $p$ is considered inconsistent if the label of $f$ obtained by segmenting the query image and the label of $p$ are not identical. 
All consistent matches are used to estimate the camera pose by applying a P3P solver~\cite{Kneip2011CVPR,Haralick94IJCV} inside a RANSAC~\cite{Fischler81CACM} loop.

\PAR{Geometric-Semantic Match Consistency (GSMC)~\cite{Toft2018ECCV}. } 
Assuming that the gravity direction and an estimate of the camera height above the ground is known, \cite{Toft2018ECCV} proposes a more complicated match consistency filter. 
For each 2D-3D correspondences, again obtained by matching image features against a SfM model, a set of camera pose hypotheses is generated. 
For each such pose, the 3D points in the model (including points that are non-matching) are projected into the query image. 
The projections are used to measure a semantic consistency score for the pose by counting the number of points projecting into a query image region with the same label as the point. 
The highest score from all poses of a match is then the semantic consistency score of that correspondence. 
The scores are normalized and used to bias RANSAC's sampling strategy to prefer selecting more semantically consistent matches.
While performing significantly better than SSMC~\cite{Toft2018ECCV}, GSMC makes additional assumptions and is computationally less efficient. 

\PAR{Particle Filter-based Semantic Localization (PFSL)~\cite{Stenborg2018LongTermVL}. } 
In this approach, localization is approached as a filtering problem where we, in addition to a sequence of camera images, also have access to noisy odometry information.
Both these sources are combined in a particle filter to sequentially estimate the pose of the camera by letting each particle describe a possible camera pose.
In the update step of the particle filter, the new weight of each particle is set proportional to how well the projection of the 3D point cloud matches the segmentation of the current image.
A 3D point $p$ is assumed to match well if the pixel which $p$ is projected to has the same label as $p$. Note that this approach does not depend on forming direct 2D-3D correspondences using, e.g., SIFT-descriptors, and is therefore more reliant on discriminative segmentation labels.

\section{Experiments}
The main focus of our experiments is evaluating the impact of using FGSNs for ``semantic" visual localization. 
In addition, we investigate whether the clusters learned by the FGSNs carry semantic information. 

\PAR{Network variations.}  
For training, we use two cross-season correspondence datasets from~\cite{larsson2019corr}, namely the CMU Seasons Correspondence Dataset and the Oxford RobotCar Correspondence Dataset.
The available samples are split into a training set (70$\%$ of the samples) and a validation set (30$\%$ of the samples). The corresponding images are geometrically separated from the query images in the Extended CMU Seasons and RobotCar Seasons benchmarks~\cite{Sattler2018CVPR} used for evaluating the localization approaches.

In addition to comparing our results to several baselines, we investigate the impact of varying the number of output clusters as well as the impact of pretraining. 
For the latter, we evaluate a first variant that initializes the base of the network with weights from a network trained on ImageNet~\cite{deng2009imagenet}, while randomly initializing the rest of the network weights. 
A second variant uses a network pre-trained for semantic segmentation using the fine annotations of the Cityscapes dataset~\cite{Cordts2016Cityscapes} and the training set of the Mapillary Vistas dataset~\cite{neuhold2017mapillary}. To be able to combine these two datasets we mapped the Vistas semantic labels to the Cityscapes labels, hence 19 semantic classes were used during training.

Further, we train FGSNs with varying number of output clusters on Cityscapes and Vistas only. For these experiments $\mathcal{L}_{corr}$ was not used since there are no available correspondences for these datasets.

\begin{table}
\begin{center}
\scriptsize{
\begin{tabular}{|cc?cc|cc|}
\hline
\multirow{2}{*}{Init} & \multirow{2}{*}{Clusters} & \multicolumn{2}{c|}{CMU} & \multicolumn{2}{c|}{RobotCar} \\ \cline{3-6}
&  & CS & WD & CS & WD \\ \hline
Seg & $20$ & $40.1$ & $33.7$ & $32.5$ & $28.0$ \\ \hline
Seg & $100$ & $47.9$ & $36.6$ & $41.5$ & $27.2$ \\ \hline
Seg & $200$ & $47.0$ & $36.6$ & $41.7$ & $32.1$ \\ \hline
Seg & $1000$ & $45.7$ & $35.8$ & $35.6$ & $26.1$ \\ \hline
Class & $200$ & $28.8$ & $26.7$ & $24.0$ & $24.7$ \\ \hline
Class & $1000$ & $18.1$ & $22.2$ & $18.4$ & $23.0$ \\ \hline
\end{tabular}
}
\end{center}
\caption{Measuring the semantic information contained in our clusters. Using models trained on the CMU or RobotCar Correspondence data, we measure the normalized mutual information (in $\%$) between our clusters and the 19 Cityscapes classes on the Cityscapes (CS) and the WildDash (WD) validation sets. 
``Seg" networks are pre-trained on semantic segmentation and ``Class" networks on classification.}
\label{res:nmi}
\end{table}
\begin{figure*}
    \centering
    \setlength\tabcolsep{1pt} 
    {\renewcommand{\arraystretch}{0.1}
    \begin{tabular}{ccc}
    \hspace{1cm}20 & 100 & 200 \\ 
    \includegraphics[height=0.25\textwidth]{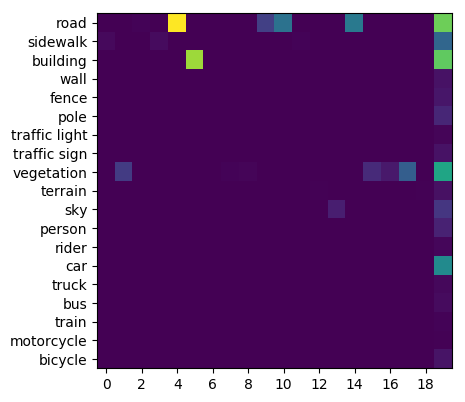} &
    \includegraphics[height=0.25\textwidth]{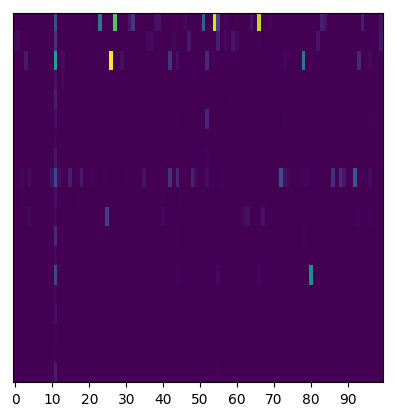} &
    \includegraphics[height=0.25\textwidth]{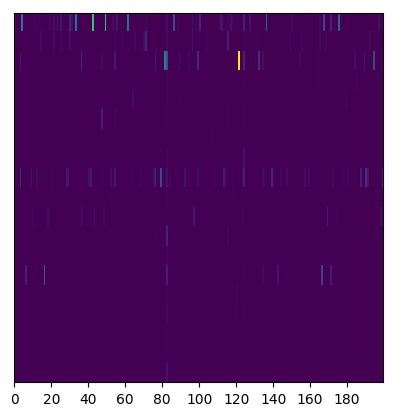} \\
    \end{tabular}
    }
    \setlength\tabcolsep{6pt} 
    \caption{Visualization of contingency tables between Cityscapes classes and cluster indices for different number of clusters. The clusters were trained on the CMU Correspondence Dataset using a model pre-trained on semantic segmentation. The colormap goes from dark blue (lowest value) to yellow (highest value). The data used is the 500 images from the Cityscapes validation set. Many of the classes common in the test images such as road, building and vegetation are split into several clusters. }
    \label{fig:matchmat}
\end{figure*}

\subsection{Semantic Information in Clusters} 
\label{sec:experiments:semantics}
Our FGSNs are inspired by the task of semantic segmentation and designed with the goal of creating more fine-grained segmentations. 
Our training procedure does not enforce that our segmentations convey semantic information. 
Still, an interesting questions is whether our clusters can be related to standard semantic classes.

%
To investigate this, we calculate the normalized mutual information (NMI) to measure the shared information between cluster assignment and the semantic labels of the annotations in the Cityscapes~\cite{Cordts2016Cityscapes} validation set. Denoting the cluster assignments as X and the semantic label assignments as Y, the normalized mutual information is given by 
\begin{equation}
\text{NMI}(X;Y) = \frac{\text{I}(X,Y)}{\sqrt{\text{H}(X)\text{H}(Y)}} \enspace ,
\end{equation}
where I is the mutual information and H the entropy. If X and Y are independent, $\text{NMI}(X;Y) = 0$. 
If one of the assignments can be predicted from the other, then all information conveyed by X is shared with Y and $\text{NMI}(X;Y) = 1$. 
In addition to the Cityscapes dataset, we also compare the cluster assignments with the same 19 classes on the WildDash dataset~\cite{zendel2018wilddash}, which is designed to evaluate the robustness of segmentation methods under a wide range of conditions. 

Tab.~\ref{res:nmi} shows the NMI for our networks. As expected, the networks pre-trained for semantic segmentation achieve a higher NMI compared to the networks pre-trained for classification. Intuitively, the clusters should thus contain semantic information that could be used for localization. However, a high NMI does not necessarily mean better localization performance. For example, a cluster containing pixels around the edges between house and sky would decrease the NMI between the cluster assignments and the semantic classes, but could be useful for localization. 

Fig.~\ref{fig:matchmat} shows contingency tables between Cityscapes classes and our cluster indices for the networks trained on CMU with semantic segmentation initialization. Each contingency table displays the interrelation between two sets of assignments of the same data by forming a two-dimensional histogram, where each dimension corresponds to one of the assignments. In our case, the dimensions correspond to the semantic class labels and cluster indices, respectively. As can be seen, there are many cluster indices that are assigned to the same pixels as the semantic class vegetation. Since the CMU images contain a significant amount of vegetation, this is both expected and could lead to more information that can be used to localize the images. Looking at the contingency table for the network with 20 clusters, we can see that the cluster with index 19 overlaps with several of the semantic classes. 
This implies that many pixels are assigned to this cluster, indicating that semantic information is lost. This is also reflected in the NMI (\cf Tab.~\ref{res:nmi}), which is lower for 20-cluster networks compared to those trained with more clusters. 

Fig.~\ref{fig:matchmat} also shows that many clusters do not directly correspond to semantic classes. 
This indicates that FGSNs deviate from the pre-trained networks used to initialize them. 

\begin{table*}[t]
\centering
\scriptsize{
 \begin{tabular}{|c| c | c | c | c || c | c | c || c | c |} 
 \hline
 \multicolumn{5}{|c||}{Training Configuration / Dataset} & \multicolumn{3}{c||}{Extended CMU Seasons} & \multicolumn{2}{c|}{RobotCar Seasons} \\ 
 \hline
   \multirow{3}*{\begin{sideways} FGSN \end{sideways}} &
   \multirow{3}*{\begin{sideways} Clusters \end{sideways}} & \multirow{3}*{\begin{sideways} Data \end{sideways}} & \multirow{3}*{\begin{sideways} $\mathcal{L}_{corr}$ \end{sideways}} & \multirow{3}*{\begin{sideways} Init \end{sideways}}   & Urban & Suburban & Park & all day & all night \\
 & &  & &  & 0.25 / 0.5 / 5 [m] & 0.25 / 0.5 / 5 [m] & 0.25 / 0.5 / 5 [m] & 0.25 / 0.5 / 5 [m] & 0.25 / 0.5 / 5 [m] \\
 &  & & &  & 2 / 5 / 10 [deg] & 2 / 5 / 10 [deg] & 2 / 5 / 10 [deg] & 2 / 5 / 10 [deg] & 2 / 5 / 10 [deg] \\[0.5ex] 
 \hline\hline
  & 19 & CS+V &  & & 71.8 / 77.1 / 83.5  & 56.0 / 61.6 / 71.6  & 32.8 / 36.9 / 46.0  & 60.1 / 92.3 / 99.2  & 8.2 / 21.0 / 35.7\\
 & 19 & CS+V+E & \checkmark &  & 75.4 / 80.7 / 87.1  & 56.3 / 62.1 / 72.0  & 35.0 / 39.4 / 49.0 & 60.3 / 92.2 / 98.9  & 8.2 / 21.2 / 35.7 \\
 & 66 & V &  &  & 75.4 / 80.6 / 87.2  & 57.1 / 62.6 / 72.3  & 34.2 / 38.3 / 47.7  & 60.3 / 92.6 / 99.2  & 8.9 / 20.3 / 36.6 \\
 & 66 & V+E & & & 65.8 / 70.4 / 77.6	& 47.7 / 52.7 / 63.7 &29.6 / 33.1 / 41.9 & 59.4 / 92.4 / 99.0 &	6.1 / 16.3 / 31.5 \\
 & 66 & V+E & \checkmark & & 66.5 / 71.2 / 78.2 & 48.1 / 53.3 / 64.1 & 29.2 / 32.7 / 42.1 & 59.7 / 91.2 / 98.3 & 7.2 / 19.6 / 36.1  \\
 
\hline
 \checkmark & 20 & CS+V+E & \checkmark & Seg & 76.3 / 81.7 / 87.6  & 59.7 / 65.7 / 75.3  & 42.9 / 47.7 / 56.6  & 57.2 / 88.7 / 96.7  & 1.9 / 6.5 / 18.9\\ 
 \checkmark & 100 & CS+V+E & \checkmark & Seg & 81.8 / 87.4 / 91.3  & 68.9 / 75.6 / 83.5  & 51.3 / 57.5 / {\bf 65.7}  & 61.1 / 93.0 / {\bf 99.9}  & 8.9 / 25.4 / 40.6\\ 
 \checkmark & 200 & CS+V+E & \checkmark & Seg & 81.0 / 86.7 / 91.1  & 67.7 / 74.8 / 82.8  & 50.8 / 57.2 / 65.0  & 61.3 / 93.2 / { 99.8}  & 9.6 / 25.9 / 44.1\\ 
 \checkmark & 1000 & CS+V+E & \checkmark & Seg & 78.0 / 84.0 / 89.2  & 62.8 / 70.7 / 79.6  & 45.1 / 51.9 / 60.9  & 60.6 / 92.4 / 99.1  & 6.5 / 17.9 / 35.7\\
\hline 
\checkmark* & 100 & CS+V+E & \checkmark & Seg & {\bf 85.3} / {\bf 91.0} / {\bf 94.6} & {\bf 69.5} / {\bf 76.4} / {\bf 83.7} & {\bf 51.4} / {\bf 57.6} / 65.5 & {\bf 61.6} / {\bf 93.5} / 99.7 & {\bf 11.0} / {\bf 28.4} / {\bf 45.2} \\
\hline
  \checkmark & 200 & CS+V &  & Seg & 75.8 / 82.4 / 88.2 & 60.7 / 68.5 / 77.4 & 42.5 / 48.5 / 57.2 & 59.9 / 92.9 / 99.4 & 4.7 / 11.4 / 26.8 \\
  \checkmark & 1000 & CS+V &  & Seg & 69.8 / 77.0 / 84.0 & 54.6 / 63.2 / 73.0 & 37.3 / 43.4 / 52.1 & 54.7 / 86.6 / 94.1 & 1.4 / 7.7 / 19.3 \\
\hline
 \checkmark & 200 & CS+V+E &  & Seg & 78.7 / 84.9 / 89.9 & 64.9 / 72.4 / 81.1 & 47.5 / 54.0 / 62.1 & 61.3 / 93.1 / 99.5 &	7.0 / 17.9 / 34.0\\
 \checkmark & 1000 & CS+V+E &  & Seg & 73.4 / 80.4 / 86.9 & 57.6 / 65.5 / 75.9	& 39.6 / 46.2 / 55.0 & 45.5 / 74.8 / 81.8	& 2.3 / 5.6 / 14.0\\
\hline
 \checkmark & 200 & CS+V+E & \checkmark & Class & 70.8 / 77.6 / 84.1  & 54.1 / 63.1 / 73.3  & 37.6 / 44.2 / 52.8  & 60.0 / 91.8 / 98.5  & 5.4 / 20.3 / 36.4 \\ 
 \checkmark & 1000 & CS+V+E & \checkmark & Class & 47.4 / 55.7 / 64.8  & 35.1 / 44.4 / 57.4  & 22.3 / 27.7 / 35.9  & 48.0 / 73.0 / 79.9  & 1.9 / 4.0 / 7.5\\ 
\hline
 & 19 & CS+V+O & \checkmark & & 69.7 / 74.6 / 81.1  & 53.2 / 58.6 / 69.0  & 31.2 / 35.2 / 44.2  & 11.8 / 17.0 / 20.7  & 0.0 / 0.0 / 0.2 \\ 
 \checkmark & 200 & CS+V+O & & Seg & 75.2 / 81.4 / 86.7 & 60.0 / 67.6 / 76.6	& 40.9 / 46.8 / 55.4 & 61.1 / 93.2 / 99.8 & 3.5 / 10.7 / 27.0 \\ 
  \checkmark & 200 & CS+V+O & \checkmark & Seg & 73.0 / 79.6 / 84.9  & 59.1 / 66.5 / 75.8  & 41.6 / 47.5 / 55.3 & 59.5 / 93.1 / 99.8  & 3.5 / 11.2 / 24.2  \\ 
 
\hline \hline
 \multicolumn{5}{|c||}{P3P RANSAC} & 65.3 / 70.1 / 77.6  & 44.5 / 49.7 / 61.5  & 27.3 / 30.6 / 39.6 & 58.4 / 88.6 / 97.1  & 3.7 / 10.7 / 23.3 \\ \hline
 \end{tabular}
 }
 \caption{Localization performance for the SSMC method with different segmentation networks on the Extended CMU Seasons dataset and the RobotCar dataset. The first column marks entries from this paper, for the entry marked with * clustering was not repeated during training. Column two indicates the number of clusters (or classes) output by the network. Note that for entries marked with 19 and 66 use the semantic classes of Cityscapes and Vistas respectively and were trained with the method presented in~\cite{larsson2019corr}. Column three details what datasets were used during training: CS (Cityscapes), V (Vistas), E (Extra \ie CMU for CMU results and RobotCar for RobotCar results), O (Other extra \ie RobotCar for CMU results and CMU for Robotcar Results). The fourth column indicates, with a \checkmark, if the correspondence loss was active during training while column five specifies the pretraining of the network (Seg for segmentation pretraining and Class for classification pretraining).}.
 \label{tab:res}
 \end{table*}

\begin{table*}[t]
\centering
\scriptsize{
 \begin{tabular}{|c || c | c | c |} 
 \hline
 Method / Setting & Urban & Suburban & Park \\
 m & 0.25 / 0.5 / 5 & 0.25 / 0.5 / 5 & 0.25 / 0.5 / 5 \\
 deg & 2 / 5 / 10 & 2 / 5 / 10 & 2 / 5 / 10 \\[0.5ex] 
 \hline\hline
SSMC (FGSN, 100 clusters, trained on CMU) & 85.3 / 91.0 /  94.6 & 69.5 / 76.4 / 83.7 & 51.4 / 57.6 / 65.5 \\
GSMC (FGSN, 200 clusters, trained on CMU) & 86.4 / 91.2 / 93.8 & {\bf 77.0} / {\bf 82.9} / 88.7  & 38.9 / 43.4 / 50.0  \\ \hline

HF-Net~\cite{sarlin2019coarse} & {\bf 89.5} / {\bf 94.2} / {\bf 97.9} & 76.5 / 82.7 / 92.7 & {\bf 57.4} / 64.4 / 80.4 \\
Asymmetric Hypercolumn Matching~\cite{germain2019sparse} & 65.7 / 82.7 / 91.0 &	66.5 / 82.6 / {\bf 92.9} & 54.3 / {\bf 71.6} / {\bf 84.1} \\
 GSMC~\cite{Toft2018ECCV} & 84.3 / 89.4 / 93.2  & 69.9 / 75.9 / 83.0  & 37.8 / 42.0 / 49.3  \\ 

City Scale Localization~\cite{Svarm2017PAMI} & 71.2 / 74.6 / 78.7 & 57.8 / 61.7 / 67.5 &34.5 / 37.0 / 42.2 \\
DenseVLAD~\cite{Torii15CVPR} & 14.7 / 36.3 / 83.9 & 5.3 / 18.7 / 73.9 & 5.2 / 19.1 / 62.0 \\
NetVLAD~\cite{Arandjelovic16CVPR} &12.2 / 31.5 / 89.8 &3.7 / 13.9 / 74.7 & 2.6 / 10.4 / 55.9 \\ \hline
\hline
PFSL (FGSN, 200 clusters, trained on CMU) & {\bf 95.3} / {\bf 99.5} / {\bf 100.0}  & {\bf 87.6} / {\bf 98.3} / 99.9  & {\bf 64.8} / {\bf 81.5} / 89.3  \\ \hline
PFSL~\cite{Stenborg2018LongTermVL} & 84.7 / 96.8 / 100.0  & 76.6 / 91.2 / {\bf 100.0}  & 39.0 / 61.2 / {\bf 95.6}  \\ \hline
  \end{tabular}
 }
  \caption{Comparison to state-of-the-art methods 
 on the Extended CMU Seasons dataset. Best results for single-shot image localization and sequential localization are marked separately.} 
 \label{tab:CMU_sota}
 \end{table*}
 
\subsection{Visual Localization}
To verify that the learned clusters, even though they are not necessarily semantic in nature, contain useful information for visual localization, we perform experiments on two datasets for long-term visual localization: RobotCar Seasons \cite{Sattler2018CVPR} and the Extended CMU Seasons dataset \cite{Sattler2018CVPR}. 

\PAR{Datasets.} 
The RobotCar Seasons dataset consists of 32,792 images  from the original RobotCar dataset~\cite{Maddern2017IJRR}. Of these, 20,862 constitute a reference sequence with publicly known reference poses. A map triangulated from sparse features observed in these images is available as a reference 3D model as an aid for structure-based localization methods. 
The reference images are all captured under a single condition while the 11,934 test images are captured under a wide variety of different conditions, including seasonal, weather, and illumination changes. 
We use a slightly different version of the RobotCar Seasons dataset, also used in~\cite{Sattler2018CVPR,larsson2019corr}, which consists of a test and training set. 
The RobotCar Correspondences Dataset that we use to train our FGSNs overlaps with the training set, but not the test set of this version of the RobotCar Seasons dataset. 


The Extended CMU Seasons dataset\footnote{Available on \url{visuallocalization.net}.} is a larger version  of the CMU Seasons dataset from~\cite{Sattler2018CVPR}, based on the CMU Visual localization dataset \cite{badino2011visual}. 
Like the RobotCar Seasons dataset, the Extended CMU Seasons dataset consists of a reference sequence with publicly known camera poses, as well as a hidden test set whose camera poses are not publicly available. The reference sequence consists of 10,338 images captured during the same day in favourable conditions. 
The test set consists of 56,613 images captured during a wide variety of conditions (sunny, snowy, autumn, \etc). The dataset covers urban, suburban, and park-like areas dominated by vegetation on both sides of the road. The latter are the most challenging parts of this dataset~\cite{Sattler2018CVPR}. 

Both datasets provide SIFT features for all test and training images. 
For SSMC and GSMC, we establish 2D-3D matches via descriptor matching~\cite{Toft2018ECCV}. 
Following~\cite{Toft2018ECCV}, the Lowe ratio test with a threshold of 0.9 is used to filter out outliers. 
P3P RANSAC is then run for 10,000 iterations to estimate the camera pose. 

\PAR{Evaluation measures.} 
We follow the evaluation protocol from~\cite{Sattler2018CVPR} and report the percentage of query images localized within $X$ meters and $Y$ degrees of the ground-truth poses, using the same thresholds as in~\cite{Sattler2018CVPR}. 

\PAR{Impact of the number of clusters.} 
In a first experiment, we evaluate the impact of the number of clusters learned by FGSNs on localization performance. 
For this experiment, we focus on the Simple Semantic Match Consistency (SSMC) and compare the performance of SSMC using FGSNs with varying numbers of clusters to the performance obtained with semantic segmentation algorithms. 
For the latter, we use networks jointly trained on Cityscapes and Vistas and on Cityscapes, Vistas, and the correspondence datasets~\cite{larsson2019corr}, using the 19 Cityscapes classes and the 66 Vistas classes. Note that entries marked with~\cite{larsson2019corr} also uses a correspondence loss similar to ours but for semantic classes. 

Table~\ref{tab:res} show the results of the experiments for the RobotCar and CMU datasets. 
As can be seen, using FGSNs trained with more than 20 clusters improves the localization performance. 
Especially under challenging conditions, \ie, night on RobotCar and Suburban and Park on CMU, the improvements obtained compared to semantic segmentations are substantial. 
Naturally, using too many clusters leads to an oversegmentation of the images and thus reduces the localization accuracy of SSMC. 
The experiments clearly show that SSMC benefits from using fine-grained segmentations, even though clusters might not necessarily correspond to standard semantic concepts. 

The reason why SSMC benefits from a larger number of clusters is that the corresponding segmentations provide a more discriminative representation of the query images and the 3D point cloud. 
This allows SSMC to filter out more wrong matches by enforcing label consistency. 
This in turn increases the inlier ratio and thus the probability that RANSAC finds the correct pose. 
Plots detailing the impact of FGSNs with different numbers of clusters on the number of inliers and the inlier ratio are provided in the supplementary material.

According to Table~\ref{tab:res}, adding the Extra dataset decreases performance, this is most likely explained by the fact that the network had to be re-implemented to produce the results. 

\PAR{Impact of pretraining FGSNs.} Entries marked with Class in column for of Table~\ref{tab:res} show results obtained when pretraining the base networks of our FGSNs on a classification rather than a semantic segmentations task. As can be seen, FGSNs pre-trained on a classification task result in a significantly lower performance compared to networks trained for semantic segmentation. 
This shows the importance of using segmentations that retain some semantic information, which is more the case for FGSNs pre-trained on semantic segmentation than for FGSNs pre-trained on classification (\cf Sec.~\ref{sec:experiments:semantics}). 

\PAR{Impact of using 2D-2D point correspondences} Results for networks trained without the additional dataset from \cite{larsson2019corr} or with the correspondence loss disabled (where the clustering still is done on feature from the CMU/RobotCar images), are shown in Table~\ref{tab:res} (row 11-14). 

As can be seen from the results, using fine-grained segmentation yields better results than using semantic classes on the Extended CMU Seasons dataset (\cf entries CS+V (19 classes) and V (66 classes). These networks however, achieve lower results than their counterparts trained with the correspondence datasets. This indicates that the correspondence loss is important for localization performance.

\PAR{Generalization abilities.} 
Table~\ref{tab:res} further show results obtained when training the FGSNs on a different dataset. 
We observe a substantial drop in performance compared to FGSNs trained on the same dataset.
This behavior is not unexpected since he 2D-2D correspondences used to train our FGSNs encourage the network to learn dataset-specific clusters. 
While the performance of FGSNs trained on another dataset is comparable to using networks trained for semantic segmentation, our results indicate that there is still significant room for improving FGSNs. 

\PAR{Repetition of clustering} Following the method developed by Caron \etal~\cite{caron2018deep} the clustering is repeated after a set number of training iterations. Interestingly, we noticed that not resetting the network actually gives slightly better performance, see entry marked with * in Table~\ref{tab:res}. We attribute this to the network, pre-trained for semantic segmentation, retains semantic information more easily without resetting. Further investigation of this is left as future work.

\PAR{Comparison with state-of-the-art methods.} 
In a final experiment, we compare SSMC, GSMC and PFSL in combination with FGSNs to state-of-the-art on the Extended CMU Seasons dataset. 

To this end, we compare against HF-Net~\cite{sarlin2019coarse}, a CNN-based hierarchical localization approach, Asymmetric Hypercolumn Matching~\cite{germain2019sparse}, an approach based on matching of hypercolumn features, DenseVLAD~\cite{Torii15CVPR}, a state-of-the-art image retrieval pipeline, and its trainable variant NetVLAD~\cite{Arandjelovic16CVPR}, City Scale Localization~\cite{Svarm2017PAMI}, a non-semantic approach based on 2D-3D matches, GSMC~\cite{Toft2018ECCV} using the semantic segmentation network from~\cite{Toft2018ECCV}, and PFSL~\cite{Stenborg2018LongTermVL} using semantic segmentation network from~\cite{larsson2019corr}.

As can be seen in Tab.~\ref{tab:CMU_sota}, using segmentations with more labels, as afforded by our FGSNs, improves localization performance closing the performance gap to the current state-of-the-art.
The results clearly validate the motivation behind FGSNs: using more segmentation labels to create more discriminative, yet still robust, representations for semantic visual localization. 

\section{Conclusion}
In this paper, we have presented Fine-Grained Segmentation Networks (FGSN), a novel type of convolutional neural networks that output dense fine-grained segmentations. Using $k$-means clustering, we can train FGSNs in a self-supervised manner, using the cluster assignments of image features as labels. This enables us to use arbitrarily many output classes without having to create annotations manually. In addition, we have used a 2D-2D correspondence dataset~\cite{larsson2019corr} to ensure that the classes are stable under seasonal changes and viewpoint variations. Through extensive experiments, we have shown that using more fine-grained segmentations, as those of our FGSNs, is beneficial for the task of semantic visual localization. 

Important future directions include further adapting  visual localization methods to a larger number of clusters to ensure that the increased level of detail of the output segmentations is properly used. In addition, it would be interesting to further work on the generalization of FGSNs, \eg, in combination with domain adaptation methods.

{\footnotesize \PAR{Acknowledgements} This work has been funded by the Swedish Research Council (grant no. 2016-04445), the Swedish Foundation for Strategic Research (Semantic Mapping and Visual Navigation for Smart Robots) and Vinnova / FFI (Perceptron, grant no. 2017-01942).}


\clearpage
\appendix

\noindent {\Large \bf Supplementary Material}

This supplementary material provides details that could not be included in the paper submission due to space limitations: 
Sec.~\ref{sec:contigency} provides details on the construction of the contingency tables used in Sec. 5.1 of the paper. 
Sec.~\ref{sec:localization} details the impact of using more fine-grained segmentations on the number of inliers and the inlier ratio in the context of visual localization (\cf lines 738 to 741 in the paper).
Finally, Sec.~\ref{sec:video} describes the contents of the videos that are provided as part of the supplementary material. 

\section{Contingency Tables}
\label{sec:contigency}

As mentioned in the main paper, a contingency table displays the interrelation between two sets of assignments of the same data by forming a two-dimensional histogram, where each dimension corresponds to one of the assignments. In our case, the dimensions corresponds to the semantic class labels and cluster indices respectively. In practice, to create the tables visualized in Fig.~\ref{fig:matchmat} of the main paper, we take the index of the output cluster from the FGSN, $c_i$, and the semantic class of the annotation, $t_i$, for each pixel in each image of the test set. For each pair $(c_i, t_i)$ we add one to value at row $t_i$ and column $c_i$. A parallel can be drawn to a confusion matrix that is a special case of a contingency table, with true assignments for rows and predicted assignments for columns.


\section{Visual Localization: Inlier counts and ratios}
\label{sec:localization}
Fig.~\ref{fig:supp:cdf_plots} shows cumulative distributions for the inlier count and inlier ratio for FGSNs with varying numbers of clusters. For this experiment, we use only the Simple Semantic Match Consistency (SSMC) approach. We compare using FGSNs to filtering with the 19 Cityscapes classes obtained from a network trained on Cityscapes, Vistas, and the correspondence datasets from~\cite{larsson2019corr}. In addition, we provide the results obtained without any semantic filtering as a baseline. 

As can be seen from Table~\ref{tab:res} of the main paper, SSMC benefits from using more fine-grained segmentations up to a certain point. For 100 and 200 clusters, the localization performance is considerably better compared to the baseline of using semantic classes. Fig.~\ref{fig:supp:cdf_plots} shows that the inlier ratio CDF is lower for these, meaning that more outliers have been removed, thus increasing the probability that RANSAC finds the correct pose. For 1000 clusters however, the segmentations become too detailed. 
This results in a high inlier ratio since many outliers are removed. 
However, it also results in a lower absolute number of inlier since also correct matches are removed. This ultimately leads to a lower localization performance.


\begin{figure*}[h!]
    \centering
  \setlength\tabcolsep{1pt} 
  \begin{tabular}{ccc}
     
      \begin{turn}{90} 
      \textbf{\hspace{1cm}Extended CMU dataset}
      \end{turn}&
      \includegraphics[width=0.40\textwidth]{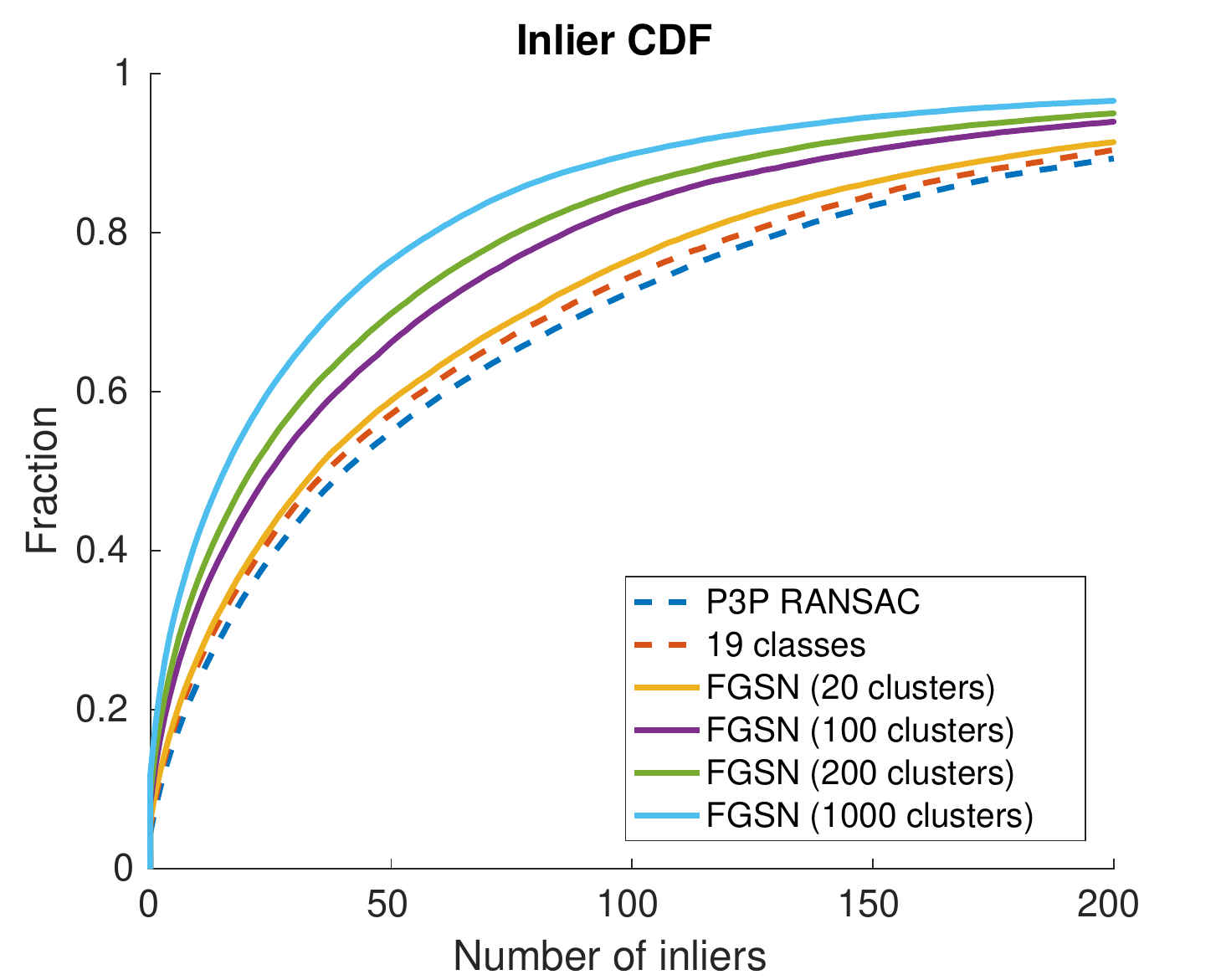}&
      \includegraphics[width=0.40\textwidth]{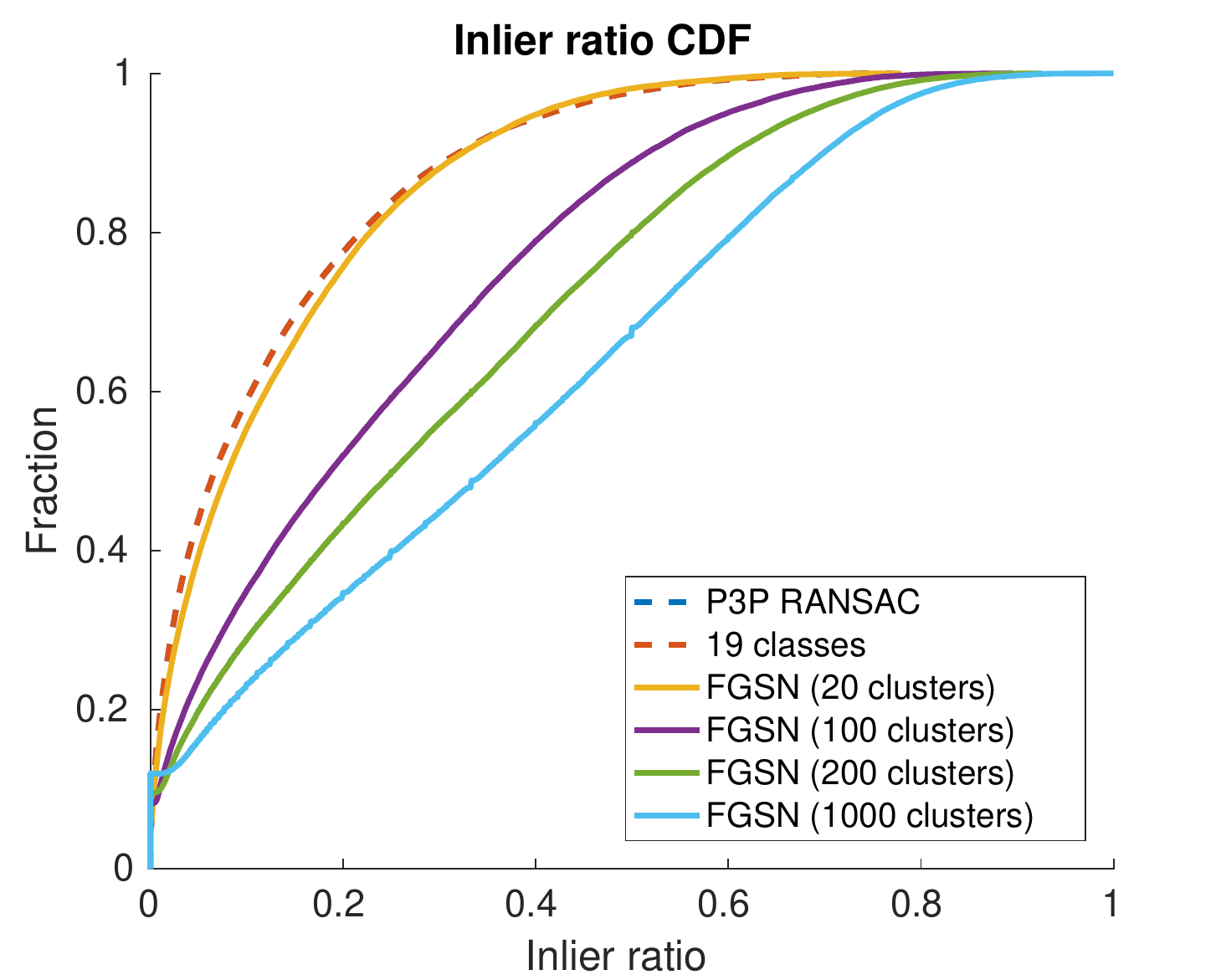} \\
      \begin{turn}{90} 
      \textbf{\hspace{1.2cm}RobotCar dataset}
      \end{turn}&

      \includegraphics[width=0.40\textwidth]{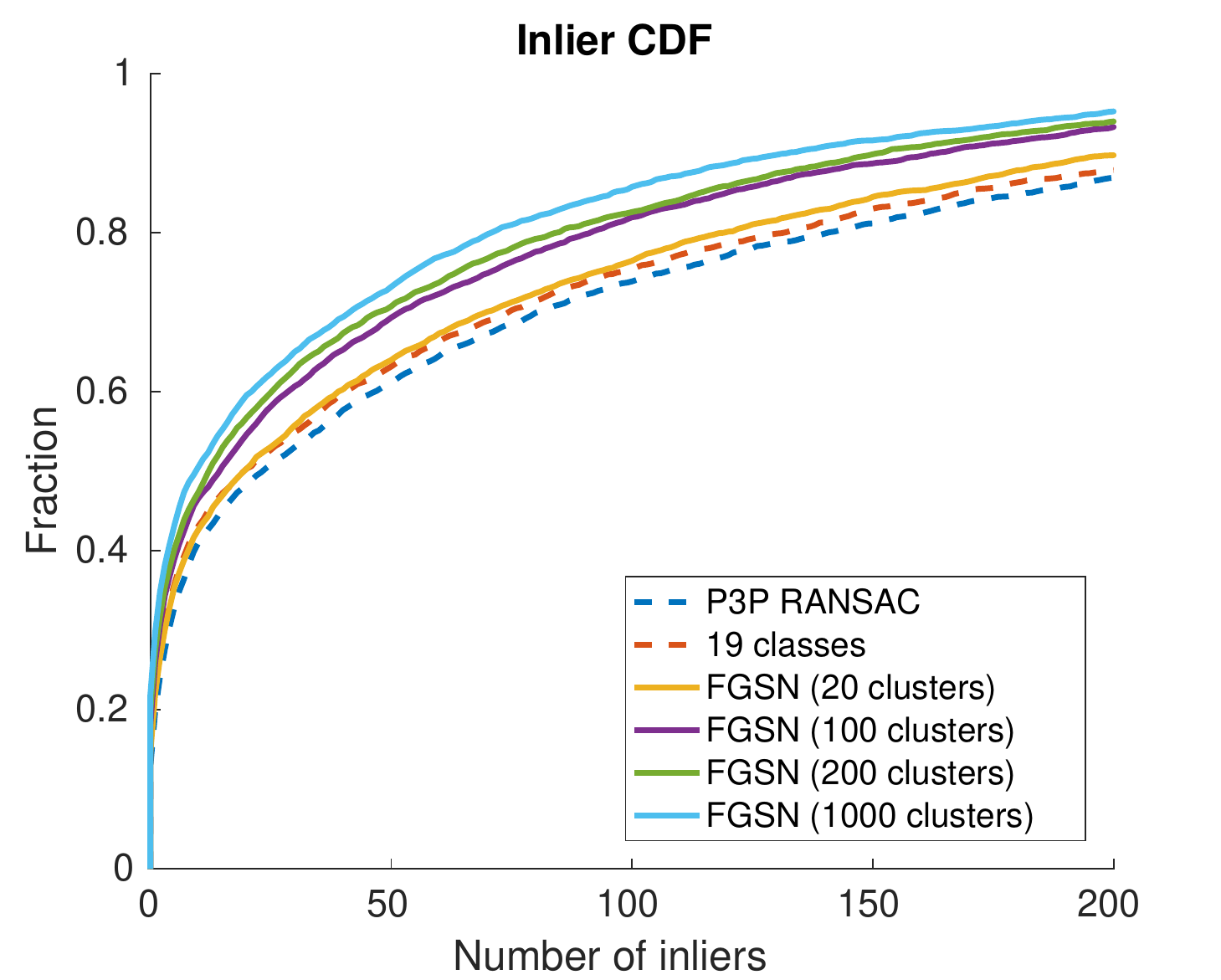}&
      \includegraphics[width=0.40\textwidth]{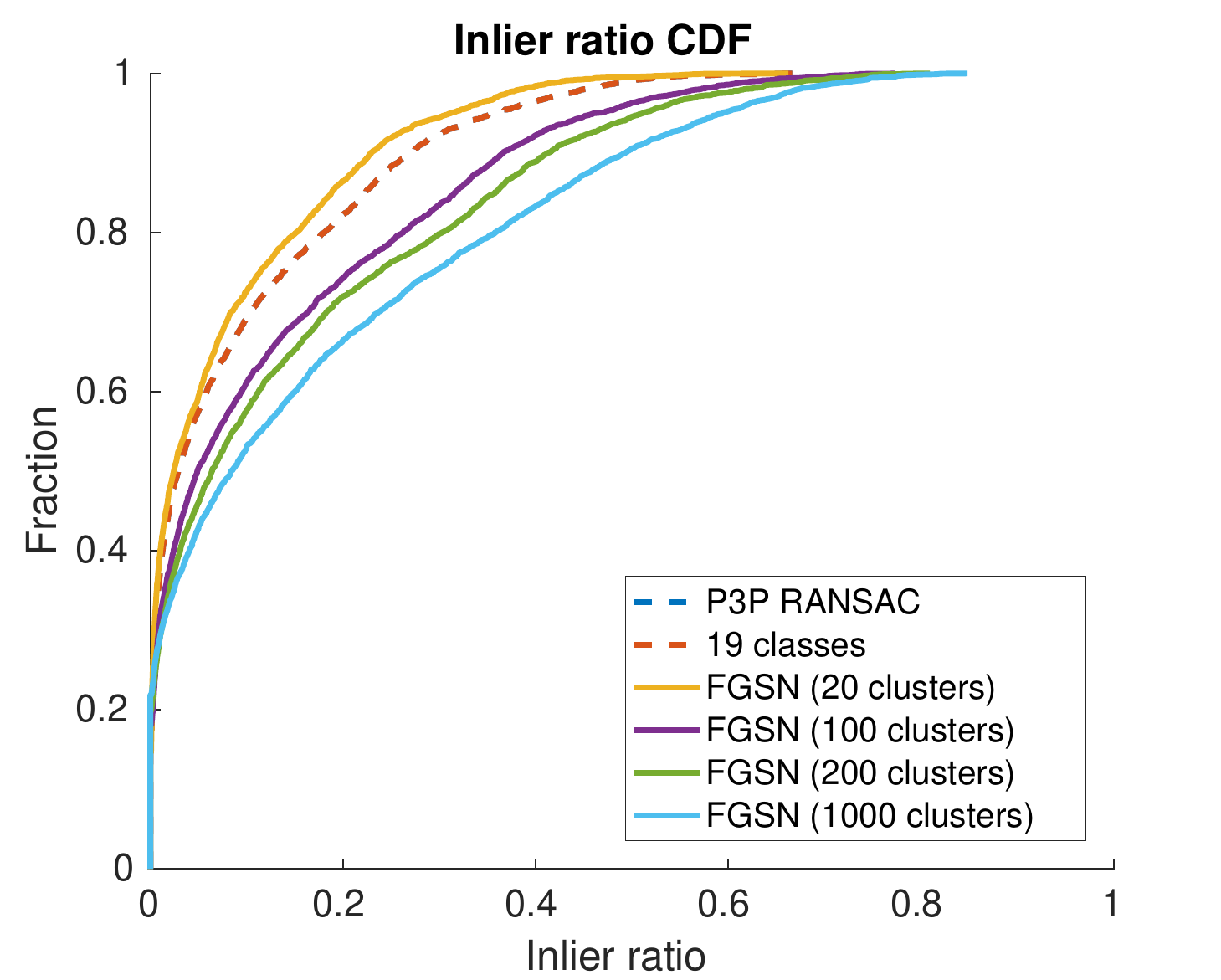}\\
      \end{tabular}
      \setlength\tabcolsep{6pt} 
      \caption{Inlier count and inlier ratio on the Extended CMU dataset (above) and the RobotCar dataset (below) using SSMC. FGSNs with varying amount of clusters are evaluated against two baselines. For the for the "19 classes"~\cite{larsson2019corr}, the Cityscapes classes are used for match consistency, while for the "P3P RANSAC" no filtering is done. Ideal curves are flat for a small number of inliers / inlier ratio and the quickly grow for a larger number of inliers / inlier ratio. \vspace{1cm}}
      \label{fig:supp:cdf_plots}
\end{figure*}

\section{Supplementary Videos}
\label{sec:video}
\subsection{Fine-Grained Segmentations}
This supplementary video contain example outputs from the FGSNs for several traversal during different seasons and image conditions. The networks used to create the segmentation were trained with correspondence loss. The video is available at \url{https://youtu.be/jXyA4wlm400}.

\subsection{Particle Filter-based Semantic Localization}
The supplementary video compares the performance of the Particle Filter-based Semantic Localization (PFSL) approach~\cite{Stenborg2018LongTermVL} when using a semantic segmentation algorithm with 19 classes trained on Cityscapes, Vistas, and the correspondence datasets from~\cite{larsson2019corr} and when using a FGSN with 200 clusters, also trained on then correspondence datasets~\cite{larsson2019corr}. For both version we use only stationary classes in the localization filter. In Cityscapes' classes that means the 11 classes "road", "sidewalk", "building", "wall", "fence", "pole" "traffic light", "traffic sign", "vegetation", "terrain", and "sky". When using FGSN we can not assign stationary classes in this way, but instead we look at which classes have many correspondences in the training data, and use those as stationary. From the training data we obtain discrete probability mass functions over the classes both for how the correspondences are distributed, $p_c(c)$, and for how all pixels in the images are distributed, $p_p(c)$. If the ratio $p_c(c)/p_p(c) > 0.2$ we select the class $c$ as stationary, and use it in the localization.

The top row shows results obtained with semantic segmentation and the bottom row shows results obtained via our FGSN. 
The left and right columns show segmentations of the left and right camera of the vehicle used to capture the CMU dataset, respectively. In addition, the points in the point cloud visible in the camera are shown in the image. Gray pixels indicate non-stationary classes or clusters and are hence not used for localization. 
The middle column shows the semantically labeled 3D point cloud of part of the extended CMU dataset (obtained by backprojecting the segmentations of the database images onto the 3D points) and the reference poses for the vehicle\footnote{The authors of \cite{Sattler2018CVPR} provided reference poses for a subset of the extended CMU dataset to aid this visualization.} (orange dots). 
The reference pose corresponding to the current images is marked with a cross. 
We also show the position estimated by PFSL (black dot) and the covariance ellipse of PFSL's estimate. The video is available at \url{https://youtu.be/-HoLNolQKoM}.

\newpage

{\small
\bibliographystyle{ieee_fullname}
\bibliography{egbib,torsten}
}

\end{document}